\pdfoutput=1

\documentclass[11pt]{article}

\usepackage[preprint]{acl}

\usepackage{times}
\usepackage{textcomp}
\usepackage{latexsym}
\usepackage{booktabs}
\usepackage{graphicx}
\usepackage{subcaption}
\usepackage{float}
\usepackage[T1]{fontenc}

\usepackage[utf8]{inputenc}

\usepackage{microtype}

\usepackage{inconsolata}

\usepackage{graphicx}
\usepackage{booktabs}
\usepackage{amsmath}
\usepackage{float}
\usepackage{xcolor}
\usepackage{tcolorbox}
\usepackage{lipsum}
\usepackage{caption}
\usepackage{soul} 
\usepackage{hyperref}
\usepackage{algorithm}
\usepackage{algpseudocode}
\usepackage{amsmath}
\usepackage{diagbox}
\usepackage{tikz}
\usepackage{authblk}
\definecolor{theme-light}{RGB}{203,227,239}    
\definecolor{theme-medium}{RGB}{108,179,217}  
\definecolor{theme-dark}{RGB}{42,120,177}  
\definecolor{bg-highlight}{RGB}{200,240,200}

\author[1]{Yuntao Shi}
\author[1]{Yi Luo}
\author[2]{Yeyun Gong}
\author[1]{Chen Lin\thanks{Corresponding author}}
\affil[1]{School of Informatics, Xiamen University}
\affil[2]{Microsoft Research Asia}

\begin{document}
\usetikzlibrary{positioning}
\title{HiCaM: A Hierarchical-Causal Modification Framework for \\ Long-Form Text Modification}
\maketitle
\newcommand{\ms}{modification suggestions}
\newcommand{\m}{modification suggestion}

\newcommand{\lftm}{long-form text modification}
\newcommand{\Lftm}{Long-form text modification}
\newcommand{\method}{HiCaM}
\begin{abstract}
Large Language Models (LLMs) have achieved remarkable success in various domains. However, when handling long-form text modification tasks, they still face two major problems: (1) producing undesired modifications by inappropriately altering or summarizing irrelevant content, and (2) missing necessary modifications to implicitly related passages that are crucial for maintaining document coherence. To address these issues, we propose HiCaM, a Hierarchical-Causal Modification framework that operates through a hierarchical summary tree and a causal graph. Furthermore, to evaluate HiCaM, we derive a multi-domain dataset from various benchmarks, providing a resource for assessing its effectiveness. Comprehensive evaluations on the dataset demonstrate significant improvements over strong LLMs, with our method achieving up to a 79.50\% win rate. These results highlight the comprehensiveness of our approach, showing consistent performance improvements across multiple models and domains.
\end{abstract}

\section{Introduction} 
The problem of \lftm{} (LTM) involves taking an original text and a modification suggestion as input, and generating a modified version that incorporates the suggested changes while preserving consistency with the original content.

\Lftm{} has significant applications across diverse domains. For example, adapting character settings in storytelling while maintaining plot coherence, changing experimental conditions in scientific protocols without introducing undesired variations, and revising academic manuscript components while maintaining logical flow. These use cases universally demand precise identification at implicitly relevant positions for synchronized updates, coupled with strategically constrained editing to prevent disruptive alterations in unrelated text regions.


\begin{figure}[H]
    \centering
    \begin{subfigure}[b]{\linewidth}
        \centering
        \includegraphics[width=\linewidth]{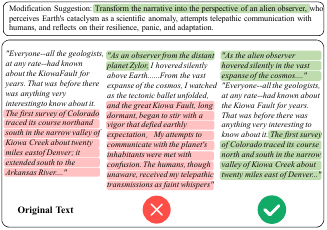}
        \caption{Undesired modifications leading to information loss and context distortion.}
        \label{fig:introcase_summary}
    \end{subfigure}
    \vspace{0.5em}
    \begin{subfigure}[b]{\linewidth}
        \centering
        \includegraphics[width=\linewidth]{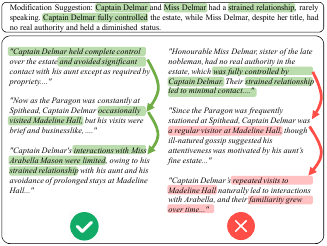}
        \caption{Missing necessary modifications for implicitly relevant content.}
        \label{fig:introcase_logic}
    \end{subfigure}
    \caption{Illustration of challenges in LTM: (a) Undesired modifications. (b) Missing necessary modifications. }
    \label{fig:introcase}
\end{figure}

In recent years, large language models (LLMs) have been applied in many fields and demonstrate exceptional capabilities. Naturally, they also hold great potential for solving the \lftm{} problem. However, as shown in Figure~\ref{fig:introcase}, they still face two significant challenges:


 \textbf{Challenge 1: Undesired modifications at inappropriate positions}, where irrelevant parts are summarized or altered, causing information loss and distorting the original completeness. Figure~\ref{fig:introcase_summary} demonstrates an example of this issue. The original text meticulously describes geographical events like the Kiowa Creek surveys through geologists'  perspectives. However, the poor modification directly blends the alien observer suggestion while discarding these key geological accounts. It reduces them to brief mentions like "the great Kiowa Fault... began to stir" without preserving the original scientific context.

 \textbf{Challenge 2: Missing necessary modifications at implicitly relevant positions}, where \ms{} fail to account for implicitly relevant parts (those not explicitly mentioned), leading to incomplete modifications that disrupt the overall coherence. As shown in Figure~\ref{fig:introcase_logic}, while both versions make the explicit modification that Captain Delmar controls the estate (Modification A), bad modifications neglect two implicit dependencies: (1) strained aunt relationship $\Rightarrow$ infrequent visits (Modification B), and (2) infrequent visits $\Rightarrow$ minimal interaction with Arabella Mason (Modification C). This \textit{partial modification} applies Modification A while ignoring B and C, preserving inconsistent Delmar-Mason interactions and violating the implicit relevance principle: modifications must propagate through all logically connected content.

For the LTM task, we propose a \textbf{Hi}erarchical-\textbf{Ca}usal \textbf{M}odification framework (\textbf{HiCaM}), which uses entity-centric analytical structures consisting of a hierarchical entity-oriented summary tree and a causal graph related to key entities. We guide LLMs to perform modifications based on these structures, resulting in outputs with both coherence and completeness.

To address Challenge 1, we introduce a \textbf{hierarchical entity-oriented summary tree} that supports hierarchical understanding and precise modification control. It ensures changes are confined to relevant regions, preserving unrelated content. The tree is constructed by recursively identifying each entity’s scope in \ms{} and summarizing its content, creating clear mappings between entities and their contexts. This entity-centric summarization divides the text into sub-sections—e.g., one for an entity’s properties, another for its functions, which can be further split into parts discussing function A and B—enabling the model to handle modifications at different granularities and avoid inappropriate changes. 

To address Challenge 2, we construct \textbf{entity-specific causal graphs} that capture logical relationships between entities throughout the text, ensuring modifications propagate properly to all related content while maintaining logical coherence. This causal representation models relationships between entities, while the hierarchical summary tree preserves relationships between sub-sections. Together, these analytical structures enable coherent and complete modifications.


To evaluate our method, we construct a multi-domain dataset by leveraging long-form texts from seven existing benchmarks. We generate \ms{} for these texts, simulating practical editing scenarios for assessment.
Experimental results demonstrate that our method consistently improves the performance of four strong LLMs on the LTM task across diverse domains. Specifically, compared with the original models, the models augmented with our method achieve win rates ranging from 56.81\% to 72.47\%, with net win rates between 13.87\% and 59.50\%. Notably, our framework on GPT-4o-mini achieves the highest average performance (68.51\% win rate, 37.50\% net win rate).



Our contributions are summarized as follows:

\begin{itemize}
    \item We introduce a novel task of \lftm{}, highlighting its importance across diverse applications. To support further research, we construct an evaluation dataset spanning multiple domains.
    \item We propose HiCaM, an approach to address \lftm{}, which leverages entity-centric analytical structures to maintain global coherence and completeness. Extensive experiments show that our approach consistently improves performance over various models and domains.
\end{itemize}


\section{Method}
\begin{figure*}[h!]
    \centering
    \includegraphics[width=\textwidth]{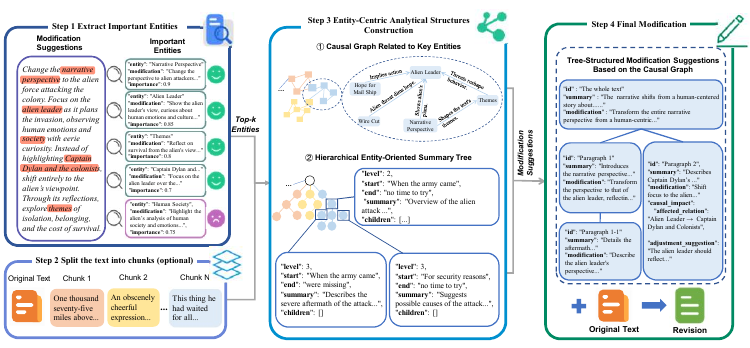}
    \caption{Overview of the \textsc{HiCaM} framework. In Step~3, the framework performs parallel processing: constructing causal graphs for key entities and building hierarchical entity-oriented summary trees, where different colors represent analytical structures in different text segments. These components jointly produce tree-structured modification suggestions, transforming the original text to the modified text while maintaining coherence.}
    \label{fig:pipeline}
\end{figure*}

\paragraph{Overview}
As illustrated in Figure~\ref{fig:pipeline}, our method begins by identifying key entities from the modification suggestions (Section~\ref{sec:entity_extract}). For overly long texts, we first divide them into manageable chunks. Next, we construct entity-centric analytical structures, including a causal graph related to key entities and a hierarchical summary tree derived from the extracted entities (Section~\ref{sec:entity-centric_analytical}). Finally, using these structures, the system generates tree-structured modification suggestions to guide the final revised output (Section~\ref{sec:final_modification}).

\subsection{Key Entity Extraction}\label{sec:entity_extract}
\begin{figure}[htbp]
    \centering
    \includegraphics[width=\linewidth]{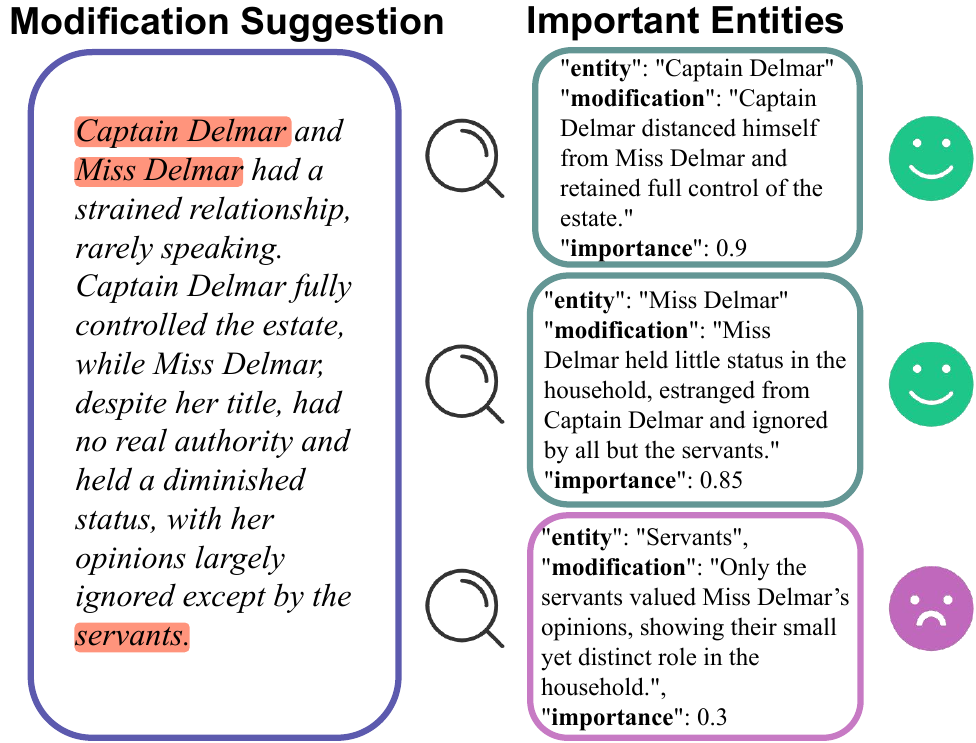}
    \caption{Example of extracted entities with importance scores and modification descriptions.}
    \label{fig:entity_example}
\end{figure}

Challenges in the modification process often arise from issues related to key entities, such as errors in entity-related logic. Based on this, we first extract the most critical entities from the provided \ms{}. For example, in Figure~\ref{fig:entity_example}, Captain Delmar (0.9) and Miss Delmar (0.8) occupy central positions in the modification suggestions, while the servants' (0.3) reactions can be inferred from the changes in the core characters' relationship. This step can be implemented in various ways, e.g., using named entity recognition models; in our method, we use a large language model (LLM) to extract the entities.

Each extracted entity is associated with three components: the entity name, its importance score, and a 'modification' field—a proposed description of how the entity should be adjusted according to the modification suggestions. The motivation behind this design is inspired by chain-of-thought (CoT)\cite{Wei2022Chain}, where the model is encouraged to generate intermediate reasoning before reaching a final decision. In practice, the model is tasked not only with identifying key entities but also with explaining what specific adjustments should be made to them based on the modification suggestions. This helps the model better identify truly important entities. It also makes the assigned scores more robust.

Given a text $T$, it can be optionally split into chunks ${C_1, \dots, C_k}$ due to its length, facilitating easier processing. The specific chunking strategy and further discussion are provided in Appendix~\ref{app:text_split}.

\subsection{Entity-Centric Analytical Structures}\label{sec:entity-centric_analytical}
Human understanding of text naturally follows a hierarchical process, where comprehension deepens as information is organized into structured layers. Inspired by this cognitive principle, we also build a hierarchical summary tree that represents the text at different granularities based on key entities. By making the relationships between sub-sections explicit, this structure helps the model avoid unnecessary rewriting and reduces the risk of undesired modifications at inappropriate positions. The tree starts from a broad representation of the text and recursively decomposes it into more specific sections, with each level capturing relationships between sub-sections at a different level of granularity. See Section~\ref{sec:summarytree} for details.

To better analyze the interactions among entities, we construct a causal graph that captures directional relationships between them. This graph reveals the relationships at the entity level, allowing the model to identify implicit dependencies and better propagate necessary modifications to relevant positions. Details are provided in Section~\ref{sec:causalgraph}.

Together, these two components form a comprehensive framework for understanding and modifying text:

A hierarchical summary tree that organizes and condenses textual information about key entities, as discussed in Section~\ref{sec:summarytree}.

A causal graph that captures directional relationships between these entities, as discussed in Section~\ref{sec:causalgraph}.

\subsubsection{Hierarchical Entity-Oriented Summary Tree}\label{sec:summarytree}

The process starts by creating a root node representing the entire text and linking it to a set of top-level segments $C_i$. Each $C_i$ is treated as a child node and is recursively processed: the model identifies meaningful sub-sections within it, identifies their scope, generates summaries focused on key entities, and creates further child nodes accordingly. This results in a hierarchical summary tree that reflects the text's structure at multiple granularities.

The recursion stops when further subdivision is unnecessary, based on the following criteria:

\noindent\textbf{\hypertarget{criteria:depth}Depth Limit}: Halts when a predefined maximum depth is reached to avoid excessive tree growth.

\noindent\textbf{\hypertarget{criteria:depth}Insufficient Segmentation Potential}: Stops if no meaningful sub-sections can be identified in the current text.

\noindent\textbf{\hypertarget{criteria:depth}Text Length Proportions}: Stops if a sub-section’s length is too close to its parent, preventing overly fine splits.

\begin{figure}[t]
\centering
\includegraphics[width=\linewidth]{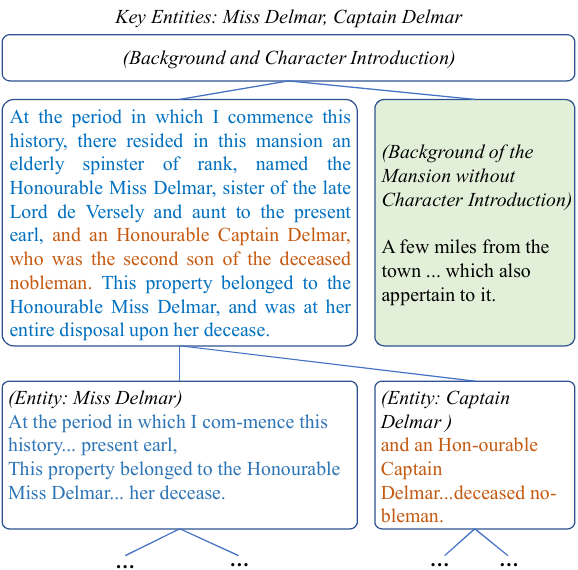}
\caption{Example of recursive processing in building a hierarchical text tree. The green portions represent background descriptions that are irrelevant to key entities, while the other sections indicate different levels of entity content obtained through recursive processing.}
\label{fig:recursive_example}
\end{figure}

Figure~\ref{fig:recursive_example} illustrates the tree-building approach. The first paragraph (green) is summarized, but since it contains no mentions of key entities, no further segmentation is performed. The second (blue) introduces key entities—Miss Delmar’s status, Captain Delmar’s lineage, and inheritance terms—requiring recursive processing. The system marks spans using opening and closing phrases (e.g., "At the period..." to "...the present earl") and summarizes each before further recursion. A detailed description of the algorithmic steps can be found in Appendix~\ref{app:summary tree construction}.

The hierarchical summary tree is a critical component for structuring the text for subsequent modifications. By organizing the text based on entities hierarchically, the tree provides a clear overview of the text’s structure. It also facilitates the application of modifications that respect both local and global contexts, ensuring that changes made to one section of the text are coherent with the overall structure and meaning.

\subsubsection{Causal Graph Related to Key Entities}\label{sec:causalgraph}
To analyze detailed interactions between entities, we construct a causal graph that explicitly models entity relationships. The graph captures:

\begin{itemize} 
    \item \textbf{Nodes}: Each node represents an entity, which is a subset of the previously extracted entities \( E \). It is identified by a unique \texttt{id} and described by a \texttt{label} that provides context about the entity's role and characteristics within the text.
    \item \textbf{Edges}: An edge from node i to node j (node i $\rightarrow$ node j) suggests that node i influences node j, with the edge's \texttt{relation} indicating the nature of the influence (e.g., "causes," "depends on," or "affects"). These edges form a network of interactions between entities.
\end{itemize}

For each chunk $C_i$, a local causal graph is first constructed. These local graphs are then merged into a unified global graph, preserving all entity relationships and causal links throughout the text. This representation enhances the hierarchical tree structure by detailing the entity-level interactions, which may span multiple sections or chunks. 

This process can be implemented in various ways, such as using relationship extraction models or employing LLMs to first extract node information and then extract edge information. In this approach, we utilize a large language model (LLM) to extract both nodes and edges directly(see Appendix~\ref{app:causal_graph_construction} for algorithmic details).

\subsection{Final Modification}\label{sec:final_modification}
After obtaining both the hierarchical summary tree and the causal graph, the final text modifications are generated through an integrated analysis of both structures.

First, tree-structured modification suggestions are generated based on the hierarchical summary tree, which serves as the backbone of the modification process. In this structure, each node explicitly specifies the required modifications for its corresponding segment (i.e., the portion of text delineated by the node’s 	\texttt{summary}, 	\texttt{start}, and 	\texttt{end} positions), ensuring that changes are systematically organized according to the text’s structural hierarchy. If a node is determined to require no modifications, it is omitted from the generated suggestions. This ensures that unmodified sections remain unchanged, preventing unnecessary alterations and preserving the structural integrity of the text.

During this modification suggestion generation process, the causal graph is actively referenced to provide a logical structure beyond what the summary tree offers. Finally, the final modification is directly derived from these tree-structured modifications, combining both the hierarchical summary tree and the causal graph to generate a coherent and complete modification.

\section{Construction of Evaluation Dataset}
To address the absence of evaluation datasets for this new task, we derive an evaluation dataset from long-form texts across multiple benchmark datasets. Specifically, we generate modification suggestions for them to simulate realistic editing scenarios, providing a structured resource for assessing the effectiveness of our proposed method.

The evaluation dataset is constructed using texts from seven benchmark datasets: NarrativeQA\cite{Kocisky2018NarrativeQA}, QuALITY\cite{Pang2022QuALITY}, GOVREPORT\cite{Huang2021Efficient}, MultiFieldQA\cite{Bai2024LongBench}, MuSiQue\cite{Trivedi20229835}, QASPER\cite{Dasigi2021Dataset}, and QMSum\cite{Zhong2021QMSum}. These datasets cover diverse domains, including fiction, government reports, academic papers, and long-context question-answering tasks. Table~\ref{tab:dataset_statistics} presents detailed statistics on text length and dataset composition, showcasing the diversity of textual sources and linguistic characteristics utilized in our evaluation.

For dataset construction, to generate \ms{} with broader impact that better challenge the model, we tailor \ms{} to each task. In summarization datasets, reference summaries are provided as meta information to help the model focus on key content and produce more targeted \ms{}. For QA datasets, queries and ground-truth answers serve as meta information, guiding deeper contextual modifications. These structured \ms{} foster meaningful refinements, creating a comprehensive evaluation testbed. Details on meta information and data construction are in Appendix~\ref{app:data_construction}.

\begin{table}[t]
\centering
\footnotesize
\setlength{\tabcolsep}{4pt}
\begin{tabular}{l c r r r}
\toprule
Dataset & Lang & Num & Max Len & Mean Len \\
\midrule
multifieldqa\_en & en & 150  & 10{,}330 & 4{,}539 \\
narrative\_qa    & en & 1548 & 10{,}853 & 5{,}929 \\
qmsum            & en & 200  & 24{,}573 & 10{,}533 \\
gov\_report\_e   & en & 310  & 22{,}423 & 6{,}457 \\
quality          & en & 381  & 6{,}004  & 4{,}158 \\
musique          & en & 200  & 6{,}866  &   964   \\
qasper           & en & 200  & 14{,}640 & 3{,}599 \\
multifieldqa\_zh & zh & 200  & 16{,}733 & 6{,}664 \\
\bottomrule
\end{tabular}
\caption{Dataset stats: lengths measured in words (English) or characters (Chinese). \texttt{data\_narrative\_qa} truncated for fair comparison.}
\label{tab:dataset_statistics}
\end{table}

\section{Experiments}

\begin{table*}[t]
\centering

\scriptsize
\begin{tabular}{p{0.48\textwidth}@{\hspace{0.01\textwidth}}p{0.48\textwidth}}
\resizebox{\linewidth}{!}{
\begin{tabular}{lccccc}
\toprule
\textbf{GPT-4o} & Win & Tie & Lose & W.R. & N.W.R. \\
\midrule
NarrativeQA & 2054 & 44 & 946 & 67.48 & 36.40 \\ 
QuALITY & 445 & 6 & 305 & 58.86 & 18.52\\
GOVREPORT & 379 & 2 & 239 & 61.13 & 22.58 \\ 
MultiFieldQA-en & 212 & 0 & 86 & 71.14 & 42.28 \\ 
MuSiQue & 239 & 2 & 135 & 63.56 & 27.66 \\ 
QASPER & 242 & 0 & 156 & 60.80 & 21.61 \\ 
QMSum & 283 & 3 & 108 & 71.83 & 44.42 \\ 
MultiFieldQA-zh & 241 & 2 & 157 & 60.25 & 21.00 \\
Avg & 511.88  & 7.38  & 266.50  & 64.38 & 29.31  \\ 
\bottomrule
\end{tabular}}
&
\resizebox{\linewidth}{!}{
\begin{tabular}{lccccc}
\toprule
\textbf{GPT-4o-mini} & Win & Tie & Lose & W.R. & N.W.R. \\
\midrule
        NarrativeQA & 1,764 & 266 & 879 & 60.64 & 30.42  \\ 
        QuALITY & 450 & 4 & 205 & 68.29 & 37.18  \\ 
        GOVREPORT & 426 & 1 & 192 & 68.82 & 37.80  \\ 
        MultiFieldQA-en & 168 & 0 & 82 & 67.20 & 34.40  \\ 
        MuSiQue & 266 & 0 & 132 & 66.83 & 33.67  \\ 
        QASPER & 245 & 0 & 105 & 70.00 & 40.00  \\ 
        QMSum & 284 & 1 & 115 & 71.00 & 42.25  \\ 
        MultiFieldQA-zh & 318 & 2 & 80 & 79.50 & 59.50  \\ 
        Avg & 490.13  & 34.25  & 223.75  & 69.03 & 39.40  \\ 
\bottomrule
\end{tabular}}
\\
\\
\resizebox{\linewidth}{!}{
\begin{tabular}{lccccc}
\toprule
\textbf{Deepseek-V3} & Win & Tie & Lose & W.R. & N.W.R. \\
\midrule
NarrativeQA & 2076 & 73 & 835 & 69.57 & 41.59 \\ 
QuALITY & 476 & 8 & 274 & 62.80 & 26.65\\
GOVREPORT & 391 & 1 & 227 & 63.17 & 26.49 \\ 
MultiFieldQA-en & 199 & 1 & 73 & 72.89 & 46.15 \\ 
MuSiQue & 274 & 2 & 122 & 68.84 & 38.19 \\ 
QASPER & 217 & 1 & 164 & 56.81 & 13.87 \\ 
QMSum & 231 & 2 & 167 & 57.75 & 16.00 \\ 
MultiFieldQA-zh & 278 & 7 & 109 & 70.56 & 42.89 \\
Avg & 517.75  & 11.88  & 246.38  & 65.30 & 31.48 \\
\bottomrule
\end{tabular}}
&
\resizebox{\linewidth}{!}{
\begin{tabular}{lccccc}
\toprule
\textbf{QWQ-32B} & Win & Tie & Lose & W.R. & N.W.R. \\
\midrule
NarrativeQA    & 2167 & 5 & 818 & 72.47 &   45.12 \\
QuALITY         & 511 &    0 &   241 & 67.95 &   35.90 \\
GOVREPORT       & 373 &    0 &   245 & 60.36 &   20.71 \\
MultiFieldQA-en & 183 &    1 &    90 & 66.79 &   33.94 \\
MuSiQue         & 245 &    0 &   155 & 61.25 &   22.50 \\
QASPER          & 260 &    1 &   122 & 67.89 &   36.03 \\
QMSum           & 245 &    1 &   153 & 61.40 &   23.06 \\
MultiFieldQA-zh	& 267 &    0 &   121 & 68.81 &   37.63 \\
Avg & 531.38  & 1.00  & 243.13  & 65.87 & 31.86\\
\bottomrule
\end{tabular}}
\end{tabular}
\vspace{0.3cm}
\\
\caption{Comparative evaluation results showing win rates (W.R.) and net win rates (N.W.R.) across four LLMs.}
\label{tab:main results}
\end{table*}

\begin{figure*}[t]
    \centering
    \begin{subfigure}{0.24\linewidth}
        \centering
        \includegraphics[width=\linewidth]{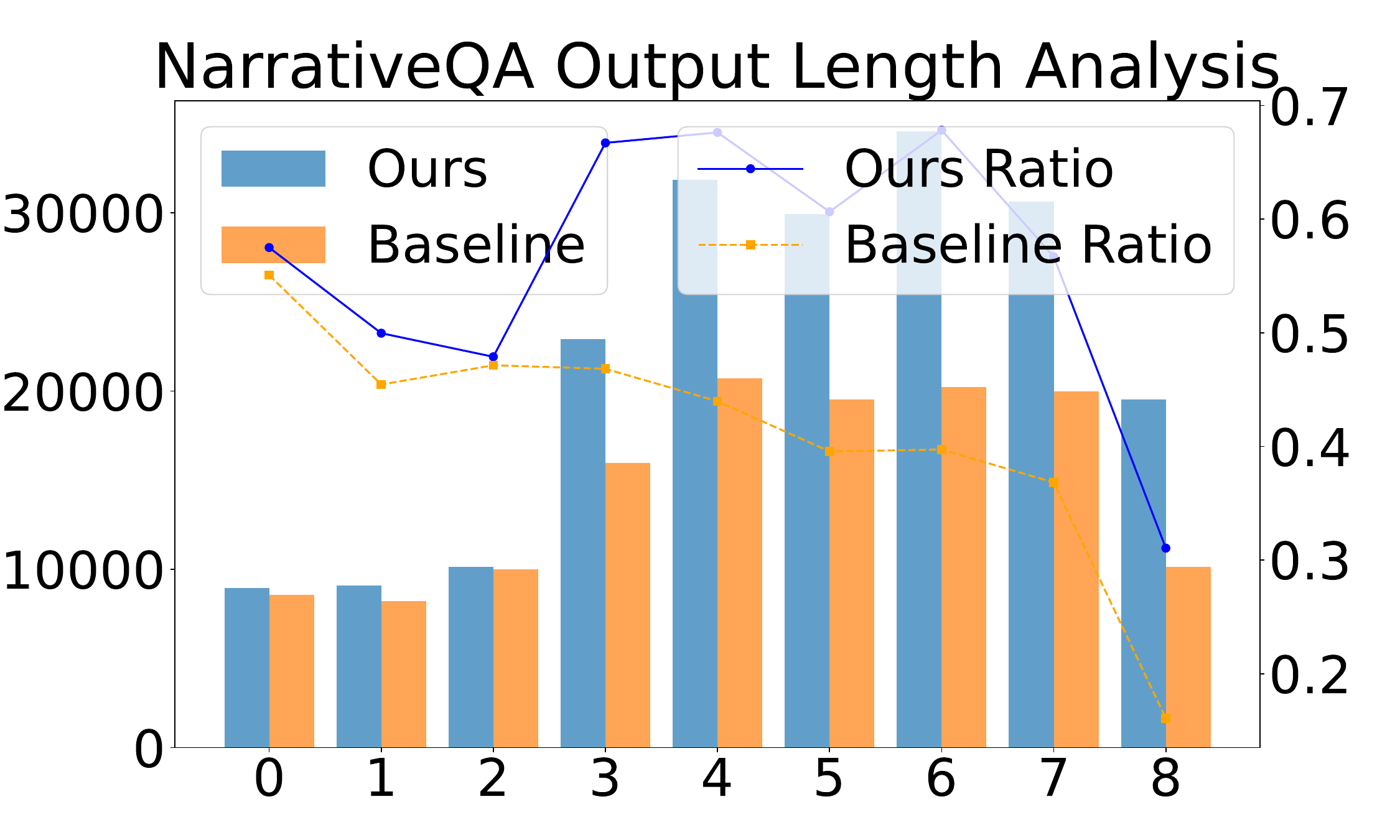}
    \end{subfigure}
    \begin{subfigure}{0.24\linewidth}
        \centering
        \includegraphics[width=\linewidth]{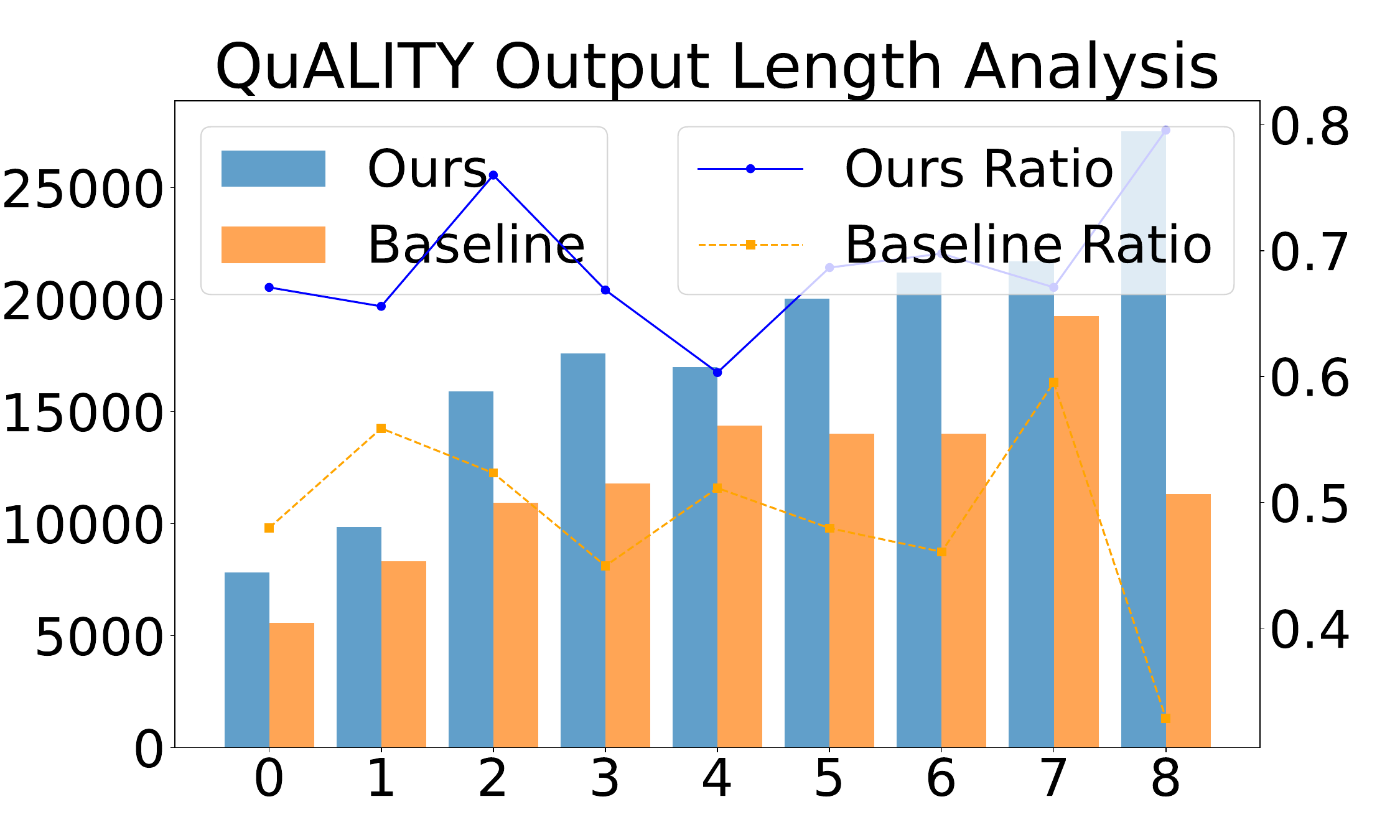}
    \end{subfigure}
    \begin{subfigure}{0.24\linewidth}
        \centering
        \includegraphics[width=\linewidth]{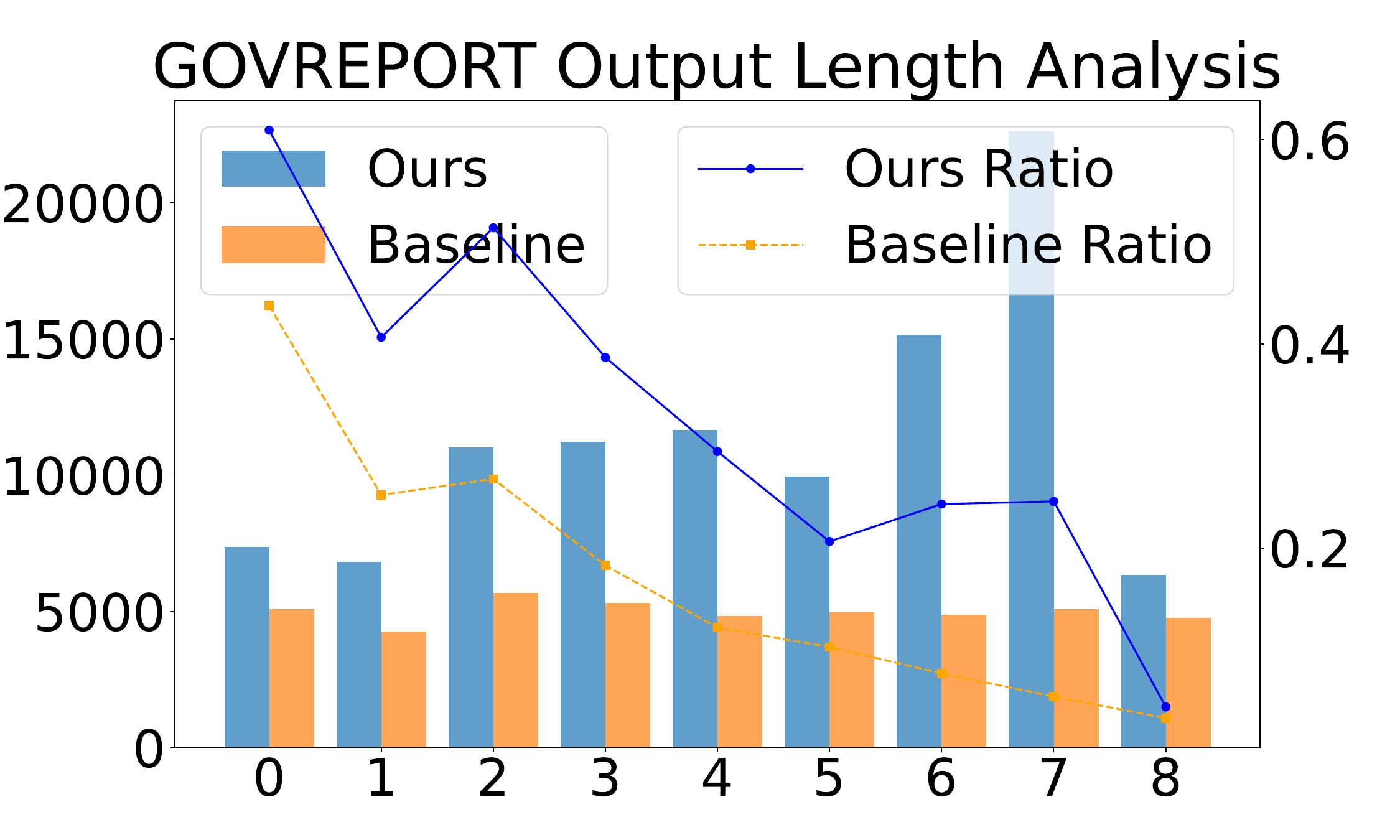}
    \end{subfigure}
    \begin{subfigure}{0.24\linewidth}
        \centering
        \includegraphics[width=\linewidth]{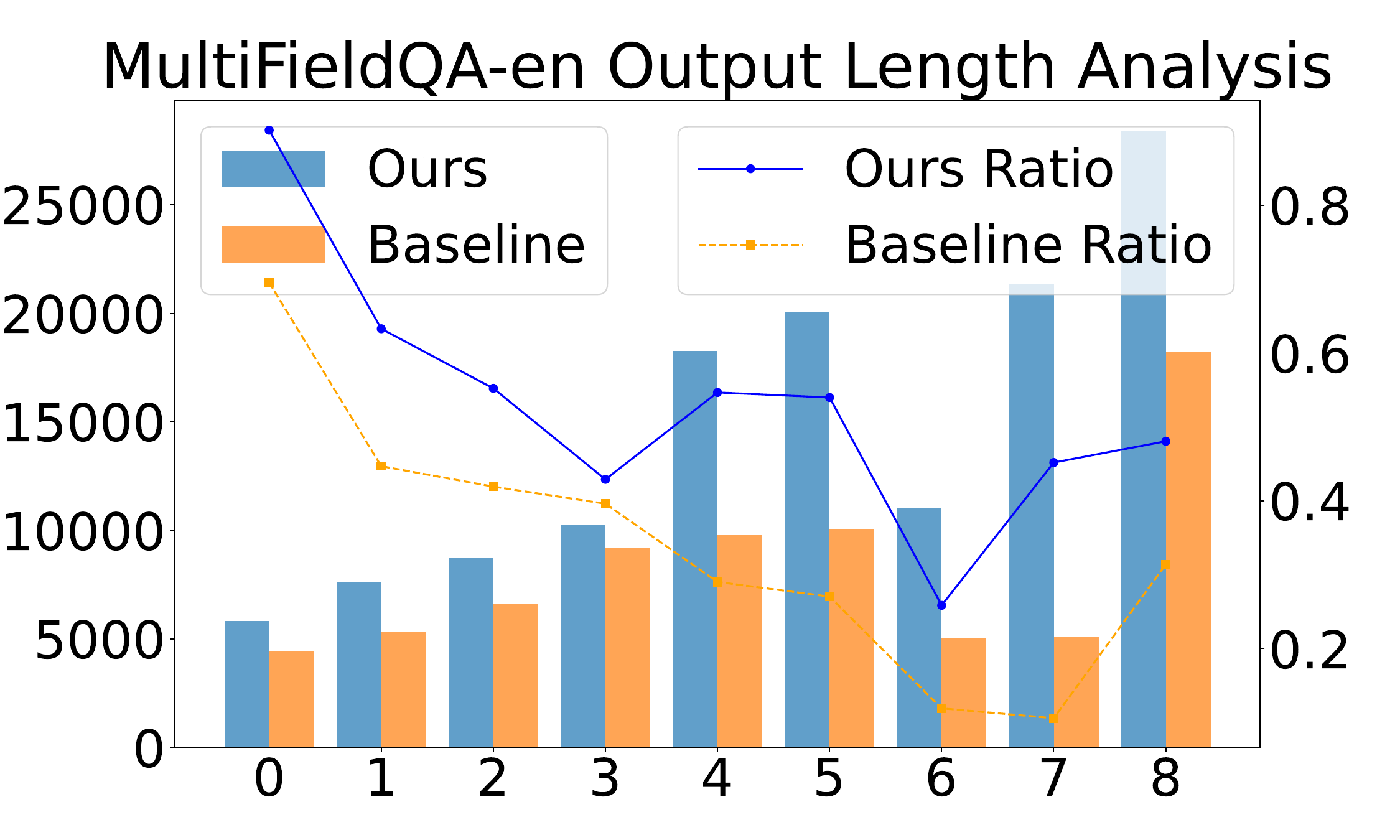}
    \end{subfigure}
    \centering
    \begin{subfigure}{0.24\linewidth}
        \centering
        \includegraphics[width=\linewidth]{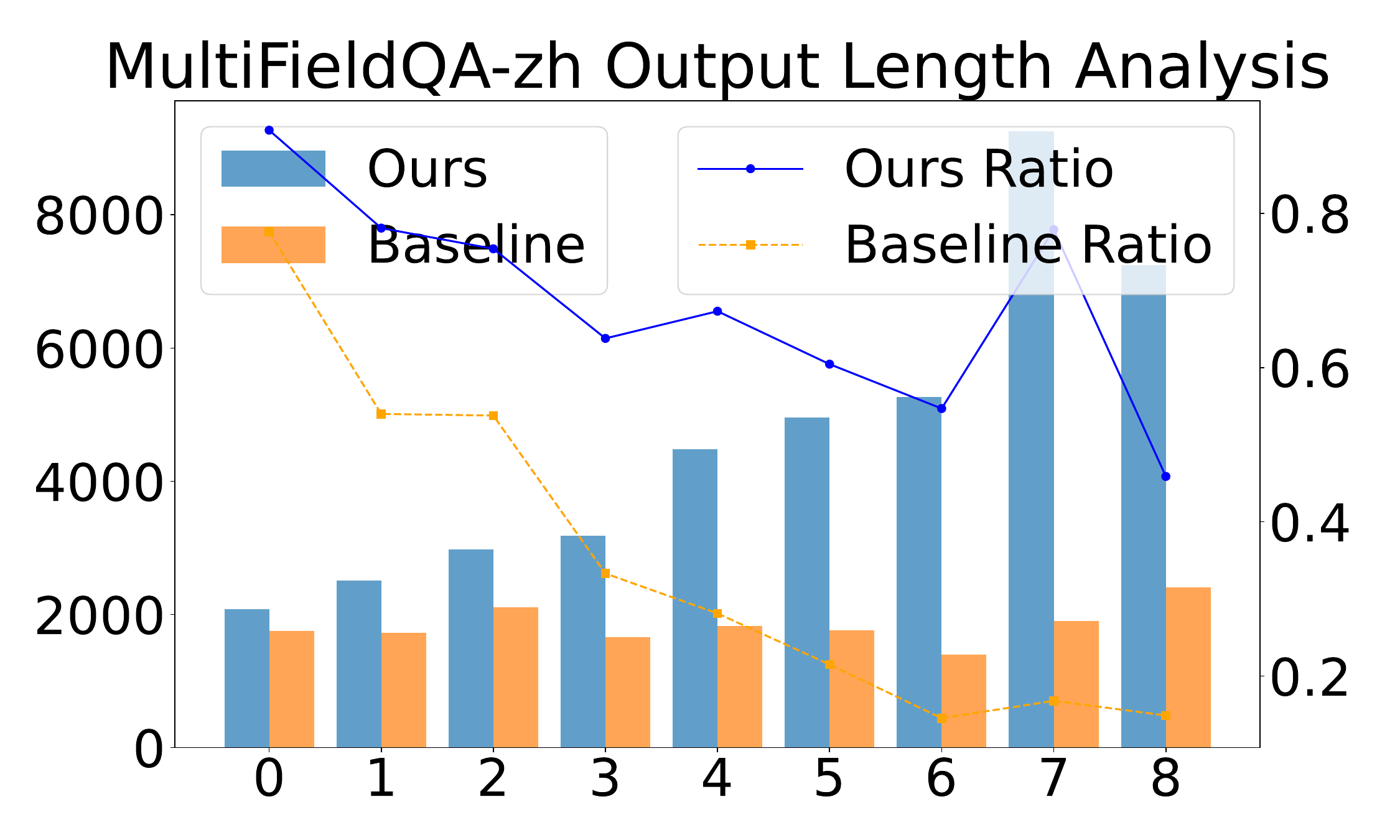}
    \end{subfigure}
    \begin{subfigure}{0.24\linewidth}
        \centering
        \includegraphics[width=\linewidth]{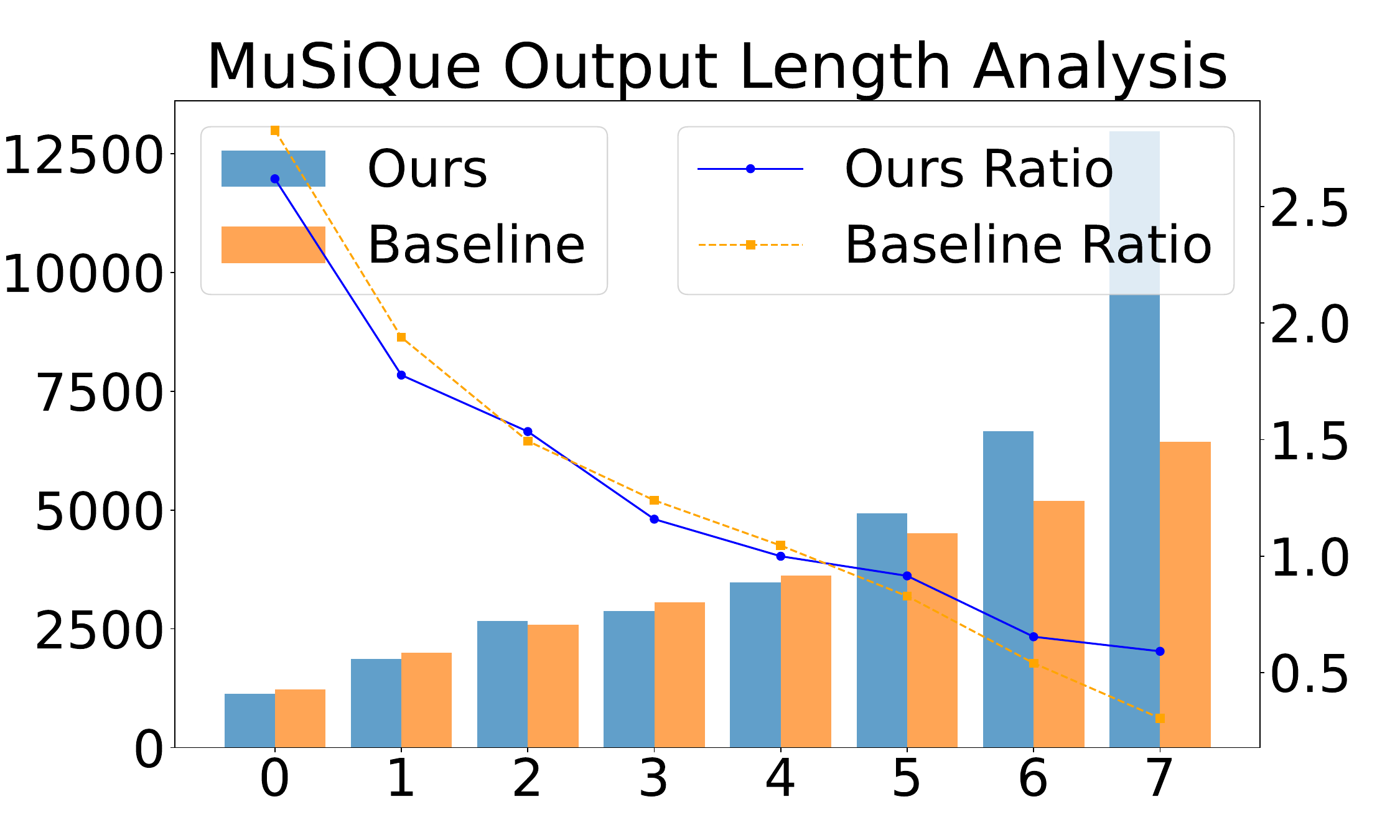}
    \end{subfigure}
    \begin{subfigure}{0.24\linewidth}
        \centering
        \includegraphics[width=\linewidth]{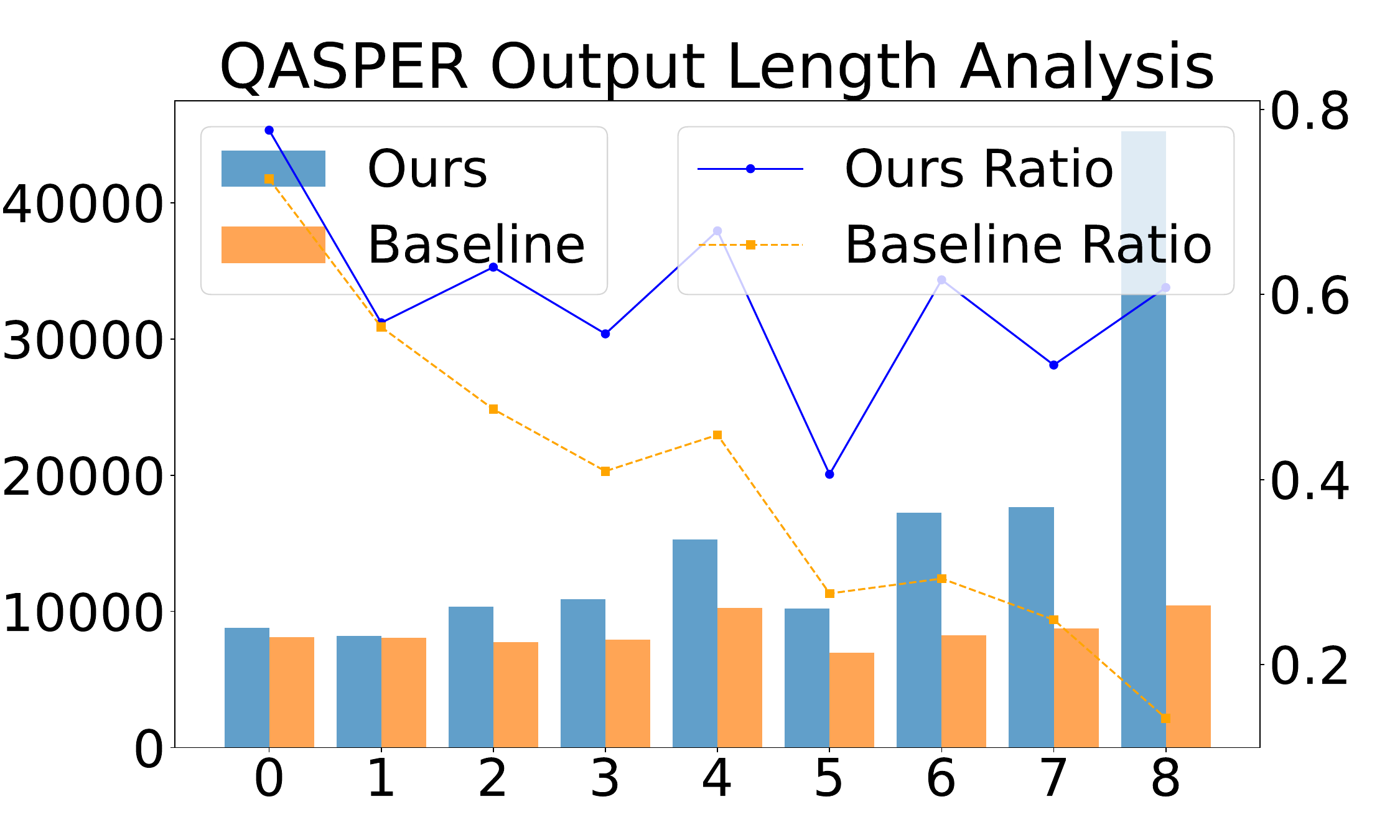}
    \end{subfigure}
    \begin{subfigure}{0.24\linewidth}
        \centering
        \includegraphics[width=\linewidth]{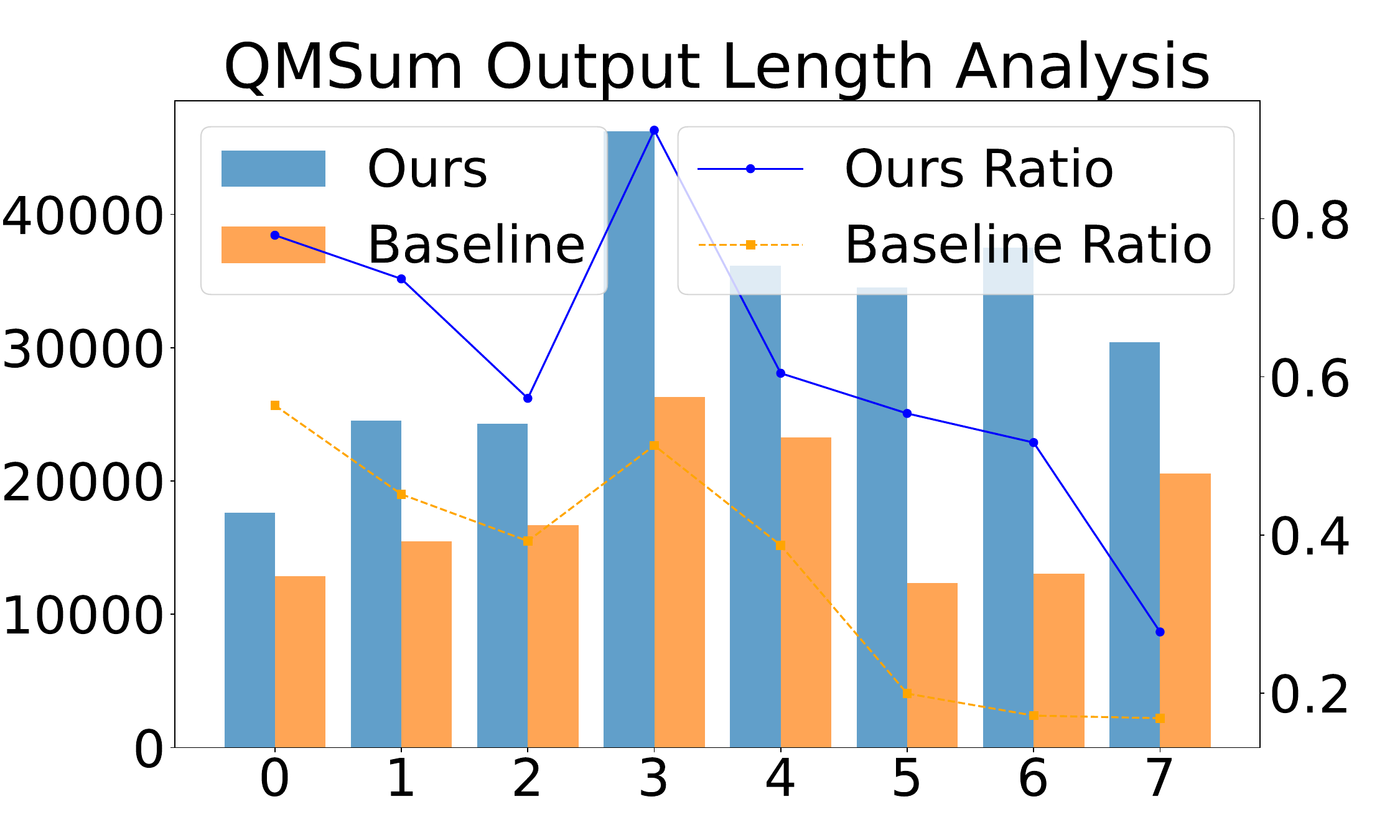}
    \end{subfigure}
    \caption{Distribution of output lengths across different datasets. The x-axis represents length bins, where outputs are grouped based on their lengths, and the y-axis shows the average output length within each bin.}
   \label{fig:length_analysis}
\end{figure*}

\subsection{Experiment Setup}
\paragraph{Implementation Details}During the pipeline, a temperature value of 0.7 is applied to the model. During evaluation, the LLMs are decoded with a temperature of 0 to minimize randomness. By default, we retrieve the top 5 entities for each input. Additionally, the depth limit for the hierarchical summary tree is set to 1, and the default input chunk size is 4096.
\paragraph{Baseline Setting} We employ several strong LLMs as baselines, including GPT-4o \cite{Hurst2024GPT}, GPT-4o-mini \cite{Hurst2024GPT}, Deepseek-V3 \cite{DeepSeekAI2024DeepSeek}, and QWQ-32B \cite{Yang2024Qwen2.5,Team2025QwQ}. For each baseline, we directly provide the original text along with the modification suggestions as input, prompting the model to output a fully modified version of the text. The prompt explicitly asks the model to ensure logical coherence and completeness in the final output.
\paragraph{Evaluation Setup}
We use GPT-4o to evaluate model outputs based on \textbf{faithfulness}, \textbf{logical coherence}, and \textbf{fluency and accessibility}, with temperature set to 0 for deterministic results. To reduce positional bias, each comparison is conducted twice by swapping candidate order, and the final win rates are aggregated. Further discussions on bias mitigation are provided in Appendix~\ref{appendix:evaluation-criteria}.In particular, experimental results and discussions on length bias are provided in Appendix~\ref{app:length_bias}.

\subsection{Main Results}
Table~\ref{tab:main results} demonstrates that our proposed framework consistently surpasses existing competitive models across all evaluated benchmarks, achieving higher win rates.
\paragraph{Models with weaker capability show larger relative gains} Among these models, GPT-4o-mini exhibits the largest net win rate of \textbf{39.40\%}. This considerable margin is likely due to its relatively weaker baseline performance, which allows more room for the framework to exert its effect. In contrast, GPT-4o, Deepseek-V3, and QWQ-32B already possess stronger initial capabilities. As a result, although their net win rates remain notable (around 30\%), the relative gains appear less striking. These models are more adept at identifying implicitly relevant positions and avoiding unnecessary changes, thereby leaving less opportunity for further enhancement by our framework.

\paragraph{Highest gains on datasets with strong internal logical structure}
As shown in Table~\ref{tab:avg_performance}, the framework achieves its highest net win rates—over \textbf{30\%}—on QuALITY, NarrativeQA, MultiFieldQA (English and Chinese), and QMSum, with MultiFieldQA-zh reaching \textbf{40.26\%}. These datasets feature texts with strong internal logical structure—narratives, police/legal documents, and meeting transcripts—where logical inconsistencies or missed relevant modifications are more likely without proper handling. In contrast, Qasper’s NLP papers lack structural elements like tables and figures, leading to lower logical complexity, and Musique articles are generally shorter, reducing the likelihood of such issues arising.

\begin{table}[t]
    \centering
    \resizebox{\columnwidth}{!}{
    \begin{tabular}{lccccc}
        \toprule
        Dataset & Win & Tie & Lose & W.R & N.W.R  \\ 
        \midrule
        NarrativeQA & 2015.25 & 97 & 869.5 & 67.54 & 38.38  \\ 
        QuALITY & 470.5 & 4.5 & 256.25 & 64.47 & 29.56  \\ 
        GOVREPORT & 392.25 & 1 & 225.75 & 63.37 & 26.90  \\ 
        MultiFieldQA-en & 190.5 & 0.5 & 82.75 & 69.51 & 39.19  \\ 
        MuSiQue & 256 & 1 & 136 & 65.12 & 30.50  \\ 
        QASPER & 241 & 0.5 & 136.75 & 63.87 & 27.88  \\ 
        QMSum & 260.75 & 1.75 & 135.75 & 65.50 & 31.43  \\ 
        MultiFieldQA-zh & 276 & 2.75 & 116.75 & 69.78 & 40.26  \\ 
        Avg & 512.78  & 13.63  & 244.94  & 66.15 & 33.01  \\  
        \bottomrule
    \end{tabular}
    }
    \caption{Average performance of models across different datasets.}
    \label{tab:avg_performance}
\end{table}

\paragraph{Analysis of Output Length}
To analyze undesired modifications, we examine the distribution of output lengths. Intuitively, if the model avoids summarizing or rewriting irrelevant content, the output length should stay close to the original. We analyze absolute output length and its ratio to the original across eight datasets, as shown in Figure~\ref{fig:length_analysis}.

Overall, two trends emerge. First, as input length increases, absolute output length also grows, but the ratio relative to the original tends to decrease. This likely results from longer inputs containing more content needing modification, prompting longer outputs. However, due to the forgetting problem in long contexts, the model may summarize or overlook details, lowering the ratio. Second, even when asked to generate a fully modified version, output length plateaus. This aligns with findings in Longwriter~\cite{BaiLongwriter}, suggesting limited exposure to long-text generation during supervised fine-tuning constrains the model's output length.

In contrast, our method shows consistent improvements in both output length and proportion as input length increases. This likely stems from the hierarchical summary tree, which helps the model grasp the text’s global structure and identify parts needing modification, reducing unnecessary summarization or rewriting.
\section{ Ablation Study and Analysis}
\begin{figure}[h]
    \centering
    \begin{minipage}[b]{0.48\linewidth}
        \includegraphics[width=\linewidth]{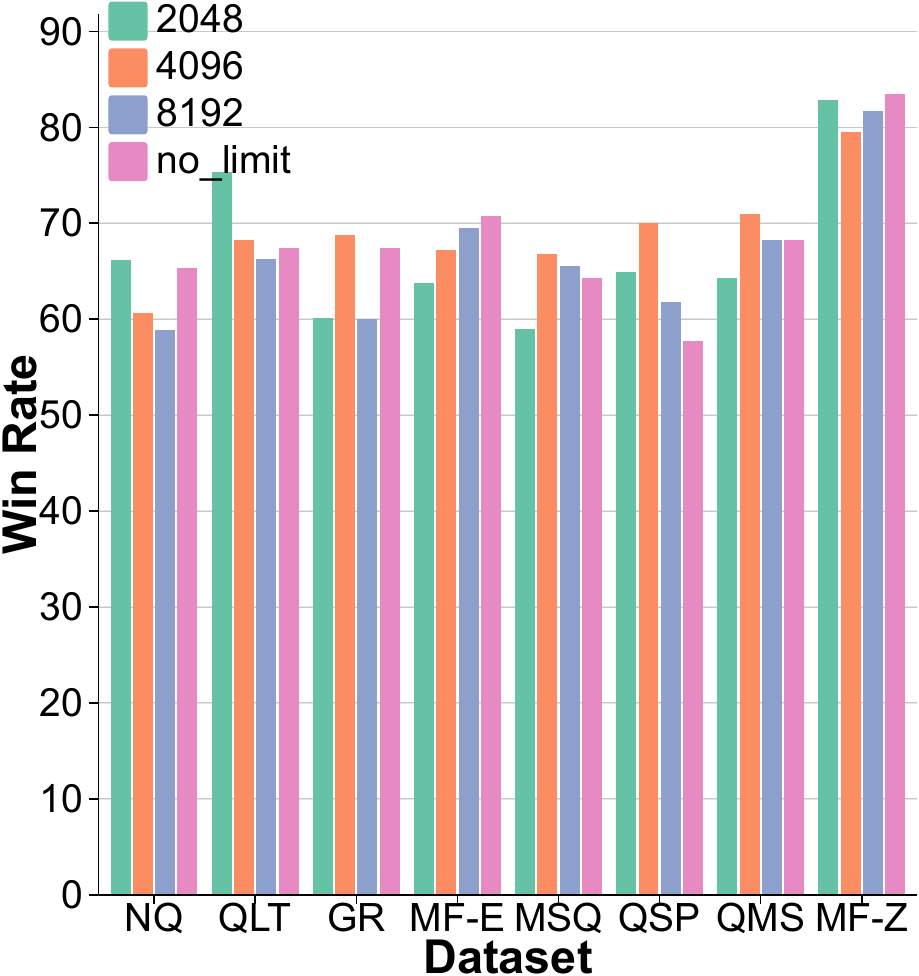}
        \caption{Win rate for different chunk sizes.}
        \label{fig:chunk_size_win_rate}
    \end{minipage}
    \hfill
    \begin{minipage}[b]{0.48\linewidth}
        \includegraphics[width=\linewidth]{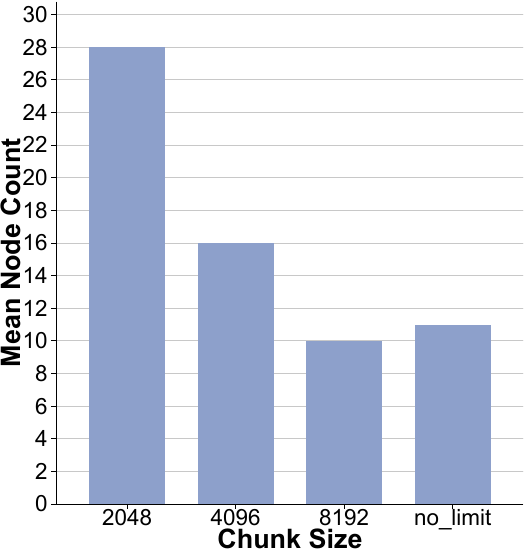}
        \caption{Tree nodes for different chunk sizes.}
        \label{fig:chunk_size_tree_nodes}
    \end{minipage}
\end{figure}

\paragraph{Effect of the Chunk Size}
Figure~\ref{fig:chunk_size_win_rate} presents the win rate for different chunk sizes, while Figure~\ref{fig:chunk_size_tree_nodes} 
illustrates the corresponding number of tree nodes. We find that different chunk sizes do not have a significant impact on the results. However, different chunk sizes have a significant impact on the number of tree nodes generated. As the chunk size increases, the number of tree nodes tends to decrease. This trend is similar to what was observed in the Graph RAG\cite{Edge2024local}, where an increase in input length causes the model to favor shorter outputs, resulting in a decline in the recall of relevant information. Conversely, a decrease in input length leads the model to prefer longer outputs, but this also introduces more redundant information and noise. Ultimately, we chose the middle value of chunk size, 4096, as the default configuration for our experiments. Chunk size results by dataset are detailed in Appendix~\ref{app:chunk_size}.

\paragraph{Effect of Causal Graph}
To evaluate the impact of the causal graph, we conduct an ablation study comparing win rates (WR) with and without causal graph integration across multiple datasets. As shown in Table \ref{tab:win_rate_ablation}, and as analyzed in the main results, datasets that require strong logical reasoning, such as MultiFieldQA-zh, QuALITY, and QMSum, exhibit significant Performance drops by an average of 20.74\%  when the causal graph is removed. In contrast, datasets like GOVREPORT and Qasper, which contain less structured logical relationships, experience a relatively smaller performance decline. This suggests that for datasets with inherently weak logical structures, enforcing causal graph extraction may have a limited impact, as the additional structural information does not significantly contribute to performance improvement. Notably, even without the causal graph, our method still outperforms direct LLM applications, as detailed in Table~\ref{tab:performance_no_causal_graph}.

\begin{table}[t]
\centering
\begin{subtable}[t]{\linewidth}
\centering
\scalebox{0.85}{
\begin{tabular}{lcccc}
\toprule
Method & NQ & QLT & GR & MF-E \\
\midrule
w/ causal graph  & 64.08 & 68.29 & 68.82 & 68.33 \\
w/o causal graph & 61.75 & 51.54 & 68.46 & 62.45 \\
\bottomrule
\end{tabular}
}
\end{subtable}

\vspace{3mm}

\begin{subtable}[t]{\linewidth}
\centering
\scalebox{0.85}{
\begin{tabular}{lcccc}
\toprule
Method & MSQ & QSP & QMS & MF-Z \\
\midrule
w/ causal graph  & 66.83 & 61.25 & 71.00 & 79.50 \\
w/o causal graph & 54.29 & 59.18 & 54.75 & 50.25 \\
\bottomrule
\end{tabular}
}
\end{subtable}
\caption{Win Rate (WR) comparison across different datasets with and without the causal graph. The datasets include NarrativeQA (NQ), QuALITY (QLT), GOVREPORT (GR), MultiFieldQA-en (MF-E), MuSiQue (MSQ), QASPER (QSP), QMSum (QMS), and MultiFieldQA-zh (MF-Z).}
\label{tab:win_rate_ablation}
\end{table}

\begin{table}[t]
    \centering
    \resizebox{\columnwidth}{!}{
    \begin{tabular}{llllll}
    \toprule
        Dataset & Win & Tie & Lose & W.R & N.W.R  \\ \midrule
        NarrativeQA & 1811 & 70 & 1052 & 61.75 & 25.88  \\ 
        QuALITY & 384 & 22 & 339 & 51.54 & 6.04  \\ 
        GOVREPORT & 421 & 2 & 192 & 68.46 & 37.24  \\ 
        MultiFieldQA-en & 153 & 0 & 92 & 62.45 & 24.90  \\ 
        MuSiQue & 215 & 2 & 179 & 54.29 & 9.09  \\ 
        QASPER & 203 & 1 & 139 & 59.18 & 18.66  \\ 
        QMSum & 219 & 2 & 179 & 54.75 & 10.00  \\ 
        MultiFieldQA-zh & 199 & 3 & 194 & 50.25 & 1.26  \\ 
        Avg & 450.63  & 12.75  & 295.75  & 57.83 & 16.63  \\ \bottomrule
    \end{tabular}
    }
        \caption{Performance without causal graph integration.}
    \label{tab:performance_no_causal_graph}
\end{table}

\paragraph{Effect of Top-k Entity Filtering}
To evaluate entity filtering, we compared different top-k selections. Without filtering, the win rate was 64.47\%. Using top-10 filtering improved it to 67.71\%, and top-5 achieved the best rate of 70.66\%. However, top-3 filtering lowered the win rate to 66.84\%, indicating that overly strict filtering may remove important entities. These results show that filtering reduces noise effectively but must balance retaining key information (see Appendix~\ref{app:topk} for details).

\section{Related Work}
Our task focuses on modifying an original long-form text according to given modification suggestions while maintaining coherence and completeness. This involves controlling text generation to meet specific modification constraints and handling long contexts since both the input and output can be lengthy. Thus, we discuss controllable text generation and long-context processing as two closely related areas essential to addressing these challenges.

\noindent\textbf{Controllable Text Generation (CTG)}
CTG focuses on steering language model outputs toward desired attributes, including content (e.g., keywords) and style (e.g., sentiment). Content control often involves architectural modifications to BERT-like models \cite{Zhang2020POINTER,He2021Parallel,Hua2020PAIR}, while style control uses either reinforcement learning with discrete rewards \cite{Zeng2024Token,Zhou2023Controlled,Dai2024Safe,Khalifa2021Distributional}, or continuous vector spaces via prompt tuning\cite{Yu2021Attribute,Senadeera2022Controlled,Wang2024Harnessing} or latent space enhancement\cite{Dathathri2020Plug,Landsman2022BeamR,Guo2024COLD}. Recent work leverages large language models for style transfer via prompting \cite{Ramirez2023Controllable,Zhang2023PCFG}, but most CTG tasks focus on coarse attributes requiring limited semantic understanding. Our work addresses the underexplored challenge of fine-grained control requiring deeper comprehension.

\noindent\textbf{Long-Context Processing}
Processing long contexts involves challenges in both input understanding and output generation. For long inputs, divide-and-conquer strategies segment text into structured units (e.g., sequential paragraphs \cite{zhao-etal-2024-longagent} or graphs \cite{Li2024GraphReader}) to maintain coherence. For long generation, methods include single-pass approaches (e.g., plan-then-generate \cite{Wang2024AutoSurvey}) and iterative refinement \cite{Quan2024Language}. Existing work typically addresses either input or output tasks separately, while our task requires joint long-context processing and generation, posing greater challenges.
\section{Conclusion}
We identify the crucial yet overlooked task of \lftm{}. We propose HiCaM - a hierarchical-causal modification framework combining multi-level text summary trees and causal graphs. This plug-and-play solution requires no additional training and can be directly integrated with existing LLMs. We introduce a dataset to evaluate \lftm{}. Extensive experiments show HiCaM achieves consistent and significant improvements over strong models across multiple domains on \lftm{}.
\section{Limitations}
Our approach has two main limitations worth noting. First, regarding computational efficiency, our framework incurs approximately 2-3 times higher computational cost compared to standard generation methods. This is because our approach requires generating multiple structured outputs during the modification process. However, we emphasize that this overhead can be effectively mitigated through parallelization strategies. Since the different structured components are generated independently, they can be processed concurrently, significantly improving practical generation speed. Furthermore, when handling longer texts that need to be processed in chunks, additional parallelization can be applied across chunks to enhance efficiency further.

Second, as shown in Table~\ref{tab:dataset_statistics}, our current evaluation is primarily conducted on English datasets (7 out of 8 datasets), with only limited testing on Chinese. While our framework is designed as a plug-and-play module that should theoretically generalize to other languages, more extensive validation across diverse languages would be needed to verify this capability fully.

\bibliography{anthology,latex/custom}

\begin{thebibliography}{35}
\providecommand{\natexlab}[1]{#1}

\bibitem[{Bai et~al.(2024)Bai, Lv, Zhang, Lyu, Tang, Huang, Du, Liu, Zeng, Hou, Dong, Tang, and Li}]{Bai2024LongBench}
Yushi Bai, Xin Lv, Jiajie Zhang, Hongchang Lyu, Jiankai Tang, Zhidian Huang, Zhengxiao Du, Xiao Liu, Aohan Zeng, Lei Hou, Yuxiao Dong, Jie Tang, and Juanzi Li. 2024.
\newblock \href {https://doi.org/10.18653/V1/2024.ACL-LONG.172} {Longbench: {A} bilingual, multitask benchmark for long context understanding}.
\newblock In \emph{Proceedings of the 62nd Annual Meeting of the Association for Computational Linguistics (Volume 1: Long Papers), {ACL} 2024, Bangkok, Thailand, August 11-16, 2024}, pages 3119--3137. Association for Computational Linguistics.

\bibitem[{Bai et~al.()Bai, Zhang, Lv, Zheng, Zhu, Hou, Dong, Tang, and Li}]{BaiLongwriter}
Yushi Bai, Jiajie Zhang, Xin Lv, Linzhi Zheng, Siqi Zhu, Lei Hou, Yuxiao Dong, Jie Tang, and Juanzi Li.
\newblock Longwriter: Unleashing 10,000+ word generation from long context llms, 2024.
\newblock \emph{URL https://arxiv. org/abs/2408.07055}.

\bibitem[{Dai et~al.(2024)Dai, Pan, Sun, Ji, Xu, Liu, Wang, and Yang}]{Dai2024Safe}
Josef Dai, Xuehai Pan, Ruiyang Sun, Jiaming Ji, Xinbo Xu, Mickel Liu, Yizhou Wang, and Yaodong Yang. 2024.
\newblock \href {https://openreview.net/forum?id=TyFrPOKYXw} {Safe {RLHF:} safe reinforcement learning from human feedback}.
\newblock In \emph{The Twelfth International Conference on Learning Representations, {ICLR} 2024, Vienna, Austria, May 7-11, 2024}. OpenReview.net.

\bibitem[{Dasigi et~al.(2021)Dasigi, Lo, Beltagy, Cohan, Smith, and Gardner}]{Dasigi2021Dataset}
Pradeep Dasigi, Kyle Lo, Iz~Beltagy, Arman Cohan, Noah~A. Smith, and Matt Gardner. 2021.
\newblock \href {https://doi.org/10.18653/V1/2021.NAACL-MAIN.365} {A dataset of information-seeking questions and answers anchored in research papers}.
\newblock In \emph{Proceedings of the 2021 Conference of the North American Chapter of the Association for Computational Linguistics: Human Language Technologies, {NAACL-HLT} 2021, Online, June 6-11, 2021}, pages 4599--4610. Association for Computational Linguistics.

\bibitem[{Dathathri et~al.(2020)Dathathri, Madotto, Lan, Hung, Frank, Molino, Yosinski, and Liu}]{Dathathri2020Plug}
Sumanth Dathathri, Andrea Madotto, Janice Lan, Jane Hung, Eric Frank, Piero Molino, Jason Yosinski, and Rosanne Liu. 2020.
\newblock \href {https://openreview.net/forum?id=H1edEyBKDS} {Plug and play language models: {A} simple approach to controlled text generation}.
\newblock In \emph{8th International Conference on Learning Representations, {ICLR} 2020, Addis Ababa, Ethiopia, April 26-30, 2020}. OpenReview.net.

\bibitem[{DeepSeek{-}AI et~al.(2024)DeepSeek{-}AI, Liu, Feng, Xue, Wang, Wu, Lu, Zhao, Deng, Zhang, Ruan, Dai, Guo, Yang, Chen, Ji, Li, Lin, Dai, Luo, Hao, Chen, Li, Zhang, Bao, Xu, Wang, Zhang, Ding, Xin, Gao, Li, Qu, Cai, Liang, Guo, Ni, Li, Wang, Chen, Chen, Yuan, Qiu, Li, Song, Dong, Hu, Gao, Guan, Huang, Yu, Wang, Zhang, Xu, Xia, Zhao, Wang, Zhang, Li, Wang, Zhang, Zhang, Tang, Li, Tian, Huang, Wang, Zhang, Wang, Zhu, Chen, Du, Chen, Jin, Ge, Zhang, Pan, Wang, Xu, Zhang, Chen, Li, Lu, Zhou, Chen, Wu, Ye, Ye, Ma, Wang, Zhou, Yu, Zhou, Pan, Wang, Yun, Pei, Sun, Xiao, and Zeng}]{DeepSeekAI2024DeepSeek}
DeepSeek{-}AI, Aixin Liu, Bei Feng, Bing Xue, Bingxuan Wang, Bochao Wu, Chengda Lu, Chenggang Zhao, Chengqi Deng, Chenyu Zhang, Chong Ruan, Damai Dai, Daya Guo, Dejian Yang, Deli Chen, Dongjie Ji, Erhang Li, Fangyun Lin, Fucong Dai, Fuli Luo, Guangbo Hao, Guanting Chen, Guowei Li, H.~Zhang, Han Bao, Hanwei Xu, Haocheng Wang, Haowei Zhang, Honghui Ding, Huajian Xin, Huazuo Gao, Hui Li, Hui Qu, J.~L. Cai, Jian Liang, Jianzhong Guo, Jiaqi Ni, Jiashi Li, Jiawei Wang, Jin Chen, Jingchang Chen, Jingyang Yuan, Junjie Qiu, Junlong Li, Junxiao Song, Kai Dong, Kai Hu, Kaige Gao, Kang Guan, Kexin Huang, Kuai Yu, Lean Wang, Lecong Zhang, Lei Xu, Leyi Xia, Liang Zhao, Litong Wang, Liyue Zhang, Meng Li, Miaojun Wang, Mingchuan Zhang, Minghua Zhang, Minghui Tang, Mingming Li, Ning Tian, Panpan Huang, Peiyi Wang, Peng Zhang, Qiancheng Wang, Qihao Zhu, Qinyu Chen, Qiushi Du, R.~J. Chen, R.~L. Jin, Ruiqi Ge, Ruisong Zhang, Ruizhe Pan, Runji Wang, Runxin Xu, Ruoyu Zhang, Ruyi Chen, S.~S. Li, Shanghao Lu, Shangyan Zhou,
  Shanhuang Chen, Shaoqing Wu, Shengfeng Ye, Shengfeng Ye, Shirong Ma, Shiyu Wang, Shuang Zhou, Shuiping Yu, Shunfeng Zhou, Shuting Pan, T.~Wang, Tao Yun, Tian Pei, Tianyu Sun, W.~L. Xiao, and Wangding Zeng. 2024.
\newblock \href {https://doi.org/10.48550/ARXIV.2412.19437} {Deepseek-v3 technical report}.
\newblock \emph{CoRR}, abs/2412.19437.

\bibitem[{Edge et~al.(2024)Edge, Trinh, Cheng, Bradley, Chao, Mody, Truitt, Metropolitansky, Ness, and Larson}]{Edge2024local}
Darren Edge, Ha~Trinh, Newman Cheng, Joshua Bradley, Alex Chao, Apurva Mody, Steven Truitt, Dasha Metropolitansky, Robert~Osazuwa Ness, and Jonathan Larson. 2024.
\newblock From local to global: A graph rag approach to query-focused summarization.
\newblock \emph{arXiv preprint arXiv:2404.16130}.

\bibitem[{Guo et~al.(2024)Guo, Yu, Zhang, Qin, and Hu}]{Guo2024COLD}
Xingang Guo, Fangxu Yu, Huan Zhang, Lianhui Qin, and Bin Hu. 2024.
\newblock \href {https://openreview.net/forum?id=yUxdk32TU6} {Cold-attack: Jailbreaking llms with stealthiness and controllability}.
\newblock In \emph{Forty-first International Conference on Machine Learning, {ICML} 2024, Vienna, Austria, July 21-27, 2024}. OpenReview.net.

\bibitem[{He(2021)}]{He2021Parallel}
Xingwei He. 2021.
\newblock \href {https://doi.org/10.18653/V1/2021.EMNLP-MAIN.681} {Parallel refinements for lexically constrained text generation with {BART}}.
\newblock In \emph{Proceedings of the 2021 Conference on Empirical Methods in Natural Language Processing, {EMNLP} 2021, Virtual Event / Punta Cana, Dominican Republic, 7-11 November, 2021}, pages 8653--8666. Association for Computational Linguistics.

\bibitem[{Hua and Wang(2020)}]{Hua2020PAIR}
Xinyu Hua and Lu~Wang. 2020.
\newblock \href {https://doi.org/10.18653/V1/2020.EMNLP-MAIN.57} {{PAIR:} planning and iterative refinement in pre-trained transformers for long text generation}.
\newblock In \emph{Proceedings of the 2020 Conference on Empirical Methods in Natural Language Processing, {EMNLP} 2020, Online, November 16-20, 2020}, pages 781--793. Association for Computational Linguistics.

\bibitem[{Huang et~al.(2021)Huang, Cao, Parulian, Ji, and Wang}]{Huang2021Efficient}
Luyang Huang, Shuyang Cao, Nikolaus Parulian, Heng Ji, and Lu~Wang. 2021.
\newblock \href {https://doi.org/10.18653/v1/2021.naacl-main.112} {Efficient attentions for long document summarization}.
\newblock In \emph{Proceedings of the 2021 Conference of the North American Chapter of the Association for Computational Linguistics: Human Language Technologies}, pages 1419--1436, Online. Association for Computational Linguistics.

\bibitem[{Hurst et~al.(2024)Hurst, Lerer, Goucher, Perelman, Ramesh, Clark, Ostrow, Welihinda, Hayes, Radford, Madry, Baker{-}Whitcomb, Beutel, Borzunov, Carney, Chow, Kirillov, Nichol, Paino, Renzin, Passos, Kirillov, Christakis, Conneau, Kamali, Jabri, Moyer, Tam, Crookes, Tootoonchian, Kumar, Vallone, Karpathy, Braunstein, Cann, Codispoti, Galu, Kondrich, Tulloch, Mishchenko, Baek, Jiang, Pelisse, Woodford, Gosalia, Dhar, Pantuliano, Nayak, Oliver, Zoph, Ghorbani, Leimberger, Rossen, Sokolowsky, Wang, Zweig, Hoover, Samic, McGrew, Spero, Giertler, Cheng, Lightcap, Walkin, Quinn, Guarraci, Hsu, Kellogg, Eastman, Lugaresi, Wainwright, Bassin, Hudson, Chu, Nelson, Li, Shern, Conger, Barette, Voss, Ding, Lu, Zhang, Beaumont, Hallacy, Koch, Gibson, Kim, Choi, McLeavey, Hesse, Fischer, Winter, Czarnecki, Jarvis, Wei, Koumouzelis, and Sherburn}]{Hurst2024GPT}
Aaron Hurst, Adam Lerer, Adam~P. Goucher, Adam Perelman, Aditya Ramesh, Aidan Clark, AJ~Ostrow, Akila Welihinda, Alan Hayes, Alec Radford, Aleksander Madry, Alex Baker{-}Whitcomb, Alex Beutel, Alex Borzunov, Alex Carney, Alex Chow, Alex Kirillov, Alex Nichol, Alex Paino, Alex Renzin, Alex~Tachard Passos, Alexander Kirillov, Alexi Christakis, Alexis Conneau, Ali Kamali, Allan Jabri, Allison Moyer, Allison Tam, Amadou Crookes, Amin Tootoonchian, Ananya Kumar, Andrea Vallone, Andrej Karpathy, Andrew Braunstein, Andrew Cann, Andrew Codispoti, Andrew Galu, Andrew Kondrich, Andrew Tulloch, Andrey Mishchenko, Angela Baek, Angela Jiang, Antoine Pelisse, Antonia Woodford, Anuj Gosalia, Arka Dhar, Ashley Pantuliano, Avi Nayak, Avital Oliver, Barret Zoph, Behrooz Ghorbani, Ben Leimberger, Ben Rossen, Ben Sokolowsky, Ben Wang, Benjamin Zweig, Beth Hoover, Blake Samic, Bob McGrew, Bobby Spero, Bogo Giertler, Bowen Cheng, Brad Lightcap, Brandon Walkin, Brendan Quinn, Brian Guarraci, Brian Hsu, Bright Kellogg, Brydon
  Eastman, Camillo Lugaresi, Carroll~L. Wainwright, Cary Bassin, Cary Hudson, Casey Chu, Chad Nelson, Chak Li, Chan~Jun Shern, Channing Conger, Charlotte Barette, Chelsea Voss, Chen Ding, Cheng Lu, Chong Zhang, Chris Beaumont, Chris Hallacy, Chris Koch, Christian Gibson, Christina Kim, Christine Choi, Christine McLeavey, Christopher Hesse, Claudia Fischer, Clemens Winter, Coley Czarnecki, Colin Jarvis, Colin Wei, Constantin Koumouzelis, and Dane Sherburn. 2024.
\newblock \href {https://doi.org/10.48550/ARXIV.2410.21276} {Gpt-4o system card}.
\newblock \emph{CoRR}, abs/2410.21276.

\bibitem[{Khalifa et~al.(2021)Khalifa, Elsahar, and Dymetman}]{Khalifa2021Distributional}
Muhammad Khalifa, Hady Elsahar, and Marc Dymetman. 2021.
\newblock \href {https://openreview.net/forum?id=jWkw45-9AbL} {A distributional approach to controlled text generation}.
\newblock In \emph{9th International Conference on Learning Representations, {ICLR} 2021, Virtual Event, Austria, May 3-7, 2021}. OpenReview.net.

\bibitem[{Kocisk{\'{y}} et~al.(2018)Kocisk{\'{y}}, Schwarz, Blunsom, Dyer, Hermann, Melis, and Grefenstette}]{Kocisky2018NarrativeQA}
Tom{\'{a}}s Kocisk{\'{y}}, Jonathan Schwarz, Phil Blunsom, Chris Dyer, Karl~Moritz Hermann, G{\'{a}}bor Melis, and Edward Grefenstette. 2018.
\newblock \href {https://doi.org/10.1162/TACL\_A\_00023} {The narrativeqa reading comprehension challenge}.
\newblock \emph{Trans. Assoc. Comput. Linguistics}, 6:317--328.

\bibitem[{Landsman et~al.(2022)Landsman, Chen, and Zaidi}]{Landsman2022BeamR}
David Landsman, Jerry~Zikun Chen, and Hussain Zaidi. 2022.
\newblock \href {https://aclanthology.org/2022.findings-aacl.40} {Beamr: Beam reweighing with attribute discriminators for controllable text generation}.
\newblock In \emph{Findings of the Association for Computational Linguistics: {AACL-IJCNLP} 2022, Online only, November 20-23, 2022}, pages 422--437. Association for Computational Linguistics.

\bibitem[{Li et~al.(2024)Li, He, Guo, Bu, Bai, Liu, Liu, Qu, Li, Ouyang, Su, and Zheng}]{Li2024GraphReader}
Shilong Li, Yancheng He, Hangyu Guo, Xingyuan Bu, Ge~Bai, Jie Liu, Jiaheng Liu, Xingwei Qu, Yangguang Li, Wanli Ouyang, Wenbo Su, and Bo~Zheng. 2024.
\newblock \href {https://aclanthology.org/2024.findings-emnlp.746} {Graphreader: Building graph-based agent to enhance long-context abilities of large language models}.
\newblock In \emph{Findings of the Association for Computational Linguistics: {EMNLP} 2024, Miami, Florida, USA, November 12-16, 2024}, pages 12758--12786. Association for Computational Linguistics.

\bibitem[{OpenAI(2023)}]{OpenAI2023GPT}
OpenAI. 2023.
\newblock \href {https://doi.org/10.48550/ARXIV.2303.08774} {{GPT-4} technical report}.
\newblock \emph{CoRR}, abs/2303.08774.

\bibitem[{Pang et~al.(2022)Pang, Parrish, Joshi, Nangia, Phang, Chen, Padmakumar, Ma, Thompson, He, and Bowman}]{Pang2022QuALITY}
Richard~Yuanzhe Pang, Alicia Parrish, Nitish Joshi, Nikita Nangia, Jason Phang, Angelica Chen, Vishakh Padmakumar, Johnny Ma, Jana Thompson, He~He, and Samuel~R. Bowman. 2022.
\newblock \href {https://doi.org/10.18653/V1/2022.NAACL-MAIN.391} {Quality: Question answering with long input texts, yes!}
\newblock In \emph{Proceedings of the 2022 Conference of the North American Chapter of the Association for Computational Linguistics: Human Language Technologies, {NAACL} 2022, Seattle, WA, United States, July 10-15, 2022}, pages 5336--5358. Association for Computational Linguistics.

\bibitem[{Quan et~al.(2024)Quan, Tang, Yu, Yang, Liu, Gao, Tu, Zhang, Zhou, and Lin}]{Quan2024Language}
Shanghaoran Quan, Tianyi Tang, Bowen Yu, An~Yang, Dayiheng Liu, Bofei Gao, Jianhong Tu, Yichang Zhang, Jingren Zhou, and Junyang Lin. 2024.
\newblock \href {https://doi.org/10.48550/ARXIV.2410.23933} {Language models can self-lengthen to generate long texts}.
\newblock \emph{CoRR}, abs/2410.23933.

\bibitem[{Ramirez et~al.(2023)Ramirez, Agarwal, Juraska, Garg, and Walker}]{Ramirez2023Controllable}
Angela Ramirez, Kartik Agarwal, Juraj Juraska, Utkarsh Garg, and Marilyn~A. Walker. 2023.
\newblock \href {https://doi.org/10.18653/V1/2023.SIGDIAL-1.32} {Controllable generation of dialogue acts for dialogue systems via few-shot response generation and ranking}.
\newblock In \emph{Proceedings of the 24th Meeting of the Special Interest Group on Discourse and Dialogue, {SIGDIAL} 2023, Prague, Czechia, September 11 - 15, 2023}, pages 355--369. Association for Computational Linguistics.

\bibitem[{Senadeera and Ive(2022)}]{Senadeera2022Controlled}
Damith~Chamalke Senadeera and Julia Ive. 2022.
\newblock \href {https://doi.org/10.48550/ARXIV.2212.02924} {Controlled text generation using {T5} based encoder-decoder soft prompt tuning and analysis of the utility of generated text in {AI}}.
\newblock \emph{CoRR}, abs/2212.02924.

\bibitem[{Team(2025)}]{Team2025QwQ}
Qwen Team. 2025.
\newblock \href {https://qwenlm.github.io/blog/qwq-32b/} {Qwq-32b: Embracing the power of reinforcement learning}.

\bibitem[{Trivedi et~al.(2022)Trivedi, Balasubramanian, Khot, and Sabharwal}]{Trivedi20229835}
Harsh Trivedi, Niranjan Balasubramanian, Tushar Khot, and Ashish Sabharwal. 2022.
\newblock \href {https://doi.org/10.1162/TACL\_A\_00475} {{\textmusicalnote} musique: Multihop questions via single-hop question composition}.
\newblock \emph{Trans. Assoc. Comput. Linguistics}, 10:539--554.

\bibitem[{Wang and Sha(2024)}]{Wang2024Harnessing}
Hao Wang and Lei Sha. 2024.
\newblock \href {https://doi.org/10.48550/ARXIV.2402.04160} {Harnessing the plug-and-play controller by prompting}.
\newblock \emph{CoRR}, abs/2402.04160.

\bibitem[{Wang et~al.(2024)Wang, Guo, Yao, Zhang, Zhang, Wu, Zhang, Dai, Zhang, Wen, Ye, Zhang, and Zhang}]{Wang2024AutoSurvey}
Yidong Wang, Qi~Guo, Wenjin Yao, Hongbo Zhang, Xin Zhang, Zhen Wu, Meishan Zhang, Xinyu Dai, Min Zhang, Qingsong Wen, Wei Ye, Shikun Zhang, and Yue Zhang. 2024.
\newblock \href {http://papers.nips.cc/paper\_files/paper/2024/hash/d07a9fc7da2e2ec0574c38d5f504d105-Abstract-Conference.html} {Autosurvey: Large language models can automatically write surveys}.
\newblock In \emph{Advances in Neural Information Processing Systems 38: Annual Conference on Neural Information Processing Systems 2024, NeurIPS 2024, Vancouver, BC, Canada, December 10 - 15, 2024}.

\bibitem[{Wei et~al.(2022)Wei, Wang, Schuurmans, Bosma, Ichter, Xia, Chi, Le, and Zhou}]{Wei2022Chain}
Jason Wei, Xuezhi Wang, Dale Schuurmans, Maarten Bosma, Brian Ichter, Fei Xia, Ed~H. Chi, Quoc~V. Le, and Denny Zhou. 2022.
\newblock \href {http://papers.nips.cc/paper\_files/paper/2022/hash/9d5609613524ecf4f15af0f7b31abca4-Abstract-Conference.html} {Chain-of-thought prompting elicits reasoning in large language models}.
\newblock In \emph{Advances in Neural Information Processing Systems 35: Annual Conference on Neural Information Processing Systems 2022, NeurIPS 2022, New Orleans, LA, USA, November 28 - December 9, 2022}.

\bibitem[{Yang et~al.(2024)Yang, Yang, Zhang, Hui, Zheng, Yu, Li, Liu, Huang, Wei, Lin, Yang, Tu, Zhang, Yang, Yang, Zhou, Lin, Dang, Lu, Bao, Yang, Yu, Li, Xue, Zhang, Zhu, Men, Lin, Li, Tang, Xia, Ren, Ren, Fan, Su, Zhang, Wan, Liu, Cui, Zhang, and Qiu}]{Yang2024Qwen2.5}
An~Yang, Baosong Yang, Beichen Zhang, Binyuan Hui, Bo~Zheng, Bowen Yu, Chengyuan Li, Dayiheng Liu, Fei Huang, Haoran Wei, Huan Lin, Jian Yang, Jianhong Tu, Jianwei Zhang, Jianxin Yang, Jiaxi Yang, Jingren Zhou, Junyang Lin, Kai Dang, Keming Lu, Keqin Bao, Kexin Yang, Le~Yu, Mei Li, Mingfeng Xue, Pei Zhang, Qin Zhu, Rui Men, Runji Lin, Tianhao Li, Tianyi Tang, Tingyu Xia, Xingzhang Ren, Xuancheng Ren, Yang Fan, Yang Su, Yichang Zhang, Yu~Wan, Yuqiong Liu, Zeyu Cui, Zhenru Zhang, and Zihan Qiu. 2024.
\newblock Qwen2.5 technical report.
\newblock \emph{arXiv preprint arXiv:2412.15115}.

\bibitem[{Yu et~al.(2021)Yu, Yu, and Sagae}]{Yu2021Attribute}
Dian Yu, Zhou Yu, and Kenji Sagae. 2021.
\newblock \href {https://doi.org/10.18653/V1/2021.FINDINGS-EMNLP.194} {Attribute alignment: Controlling text generation from pre-trained language models}.
\newblock In \emph{Findings of the Association for Computational Linguistics: {EMNLP} 2021, Virtual Event / Punta Cana, Dominican Republic, 16-20 November, 2021}, pages 2251--2268. Association for Computational Linguistics.

\bibitem[{Zeng et~al.(2024)Zeng, Liu, Ma, Yang, Zhang, and Wang}]{Zeng2024Token}
Yongcheng Zeng, Guoqing Liu, Weiyu Ma, Ning Yang, Haifeng Zhang, and Jun Wang. 2024.
\newblock \href {https://openreview.net/forum?id=1RZKuvqYCR} {Token-level direct preference optimization}.
\newblock In \emph{Forty-first International Conference on Machine Learning, {ICML} 2024, Vienna, Austria, July 21-27, 2024}. OpenReview.net.

\bibitem[{Zhang et~al.(2023)Zhang, Glass, and He}]{Zhang2023PCFG}
Jingyu Zhang, James~R. Glass, and Tianxing He. 2023.
\newblock \href {https://doi.org/10.18653/V1/2023.STARSEM-1.27} {Pcfg-based natural language interface improves generalization for controlled text generation}.
\newblock In \emph{Proceedings of the The 12th Joint Conference on Lexical and Computational Semantics, *SEM@ACL 2023, Toronto, Canada, July 13-14, 2023}, pages 295--313. Association for Computational Linguistics.

\bibitem[{Zhang et~al.(2020)Zhang, Wang, Li, Gan, Brockett, and Dolan}]{Zhang2020POINTER}
Yizhe Zhang, Guoyin Wang, Chunyuan Li, Zhe Gan, Chris Brockett, and Bill Dolan. 2020.
\newblock \href {https://doi.org/10.18653/V1/2020.EMNLP-MAIN.698} {{POINTER:} constrained progressive text generation via insertion-based generative pre-training}.
\newblock In \emph{Proceedings of the 2020 Conference on Empirical Methods in Natural Language Processing, {EMNLP} 2020, Online, November 16-20, 2020}, pages 8649--8670. Association for Computational Linguistics.

\bibitem[{Zhao et~al.(2024)Zhao, Zu, Hao, Lu, He, Ding, Gui, Zhang, and Huang}]{zhao-etal-2024-longagent}
Jun Zhao, Can Zu, Xu~Hao, Yi~Lu, Wei He, Yiwen Ding, Tao Gui, Qi~Zhang, and Xuanjing Huang. 2024.
\newblock \href {https://doi.org/10.18653/v1/2024.emnlp-main.912} {{LONGAGENT}: Achieving question answering for 128k-token-long documents through multi-agent collaboration}.
\newblock In \emph{Proceedings of the 2024 Conference on Empirical Methods in Natural Language Processing}, pages 16310--16324, Miami, Florida, USA. Association for Computational Linguistics.

\bibitem[{Zheng et~al.(2023)Zheng, Chiang, Sheng, Zhuang, Wu, Zhuang, Lin, Li, Li, Xing, Zhang, Gonzalez, and Stoica}]{Zheng2023Judging}
Lianmin Zheng, Wei{-}Lin Chiang, Ying Sheng, Siyuan Zhuang, Zhanghao Wu, Yonghao Zhuang, Zi~Lin, Zhuohan Li, Dacheng Li, Eric~P. Xing, Hao Zhang, Joseph~E. Gonzalez, and Ion Stoica. 2023.
\newblock \href {http://papers.nips.cc/paper\_files/paper/2023/hash/91f18a1287b398d378ef22505bf41832-Abstract-Datasets\_and\_Benchmarks.html} {Judging llm-as-a-judge with mt-bench and chatbot arena}.
\newblock In \emph{Advances in Neural Information Processing Systems 36: Annual Conference on Neural Information Processing Systems 2023, NeurIPS 2023, New Orleans, LA, USA, December 10 - 16, 2023}.

\bibitem[{Zhong et~al.(2021)Zhong, Yin, Yu, Zaidi, Mutuma, Jha, Awadallah, Celikyilmaz, Liu, Qiu, and Radev}]{Zhong2021QMSum}
Ming Zhong, Da~Yin, Tao Yu, Ahmad Zaidi, Mutethia Mutuma, Rahul Jha, Ahmed~Hassan Awadallah, Asli Celikyilmaz, Yang Liu, Xipeng Qiu, and Dragomir~R. Radev. 2021.
\newblock \href {https://doi.org/10.18653/V1/2021.NAACL-MAIN.472} {Qmsum: {A} new benchmark for query-based multi-domain meeting summarization}.
\newblock In \emph{Proceedings of the 2021 Conference of the North American Chapter of the Association for Computational Linguistics: Human Language Technologies, {NAACL-HLT} 2021, Online, June 6-11, 2021}, pages 5905--5921. Association for Computational Linguistics.

\bibitem[{Zhou et~al.(2023)Zhou, Jiang, Wilcox, Cotterell, and Sachan}]{Zhou2023Controlled}
Wangchunshu Zhou, Yuchen~Eleanor Jiang, Ethan Wilcox, Ryan Cotterell, and Mrinmaya Sachan. 2023.
\newblock \href {https://proceedings.mlr.press/v202/zhou23g.html} {Controlled text generation with natural language instructions}.
\newblock In \emph{International Conference on Machine Learning, {ICML} 2023, 23-29 July 2023, Honolulu, Hawaii, {USA}}, volume 202 of \emph{Proceedings of Machine Learning Research}, pages 42602--42613. {PMLR}.

\end{thebibliography}

\appendix
\newpage
\section{Algorithmic Framework}
\subsection{Algorithm for Hierarchical Summary Tree Construction}\label{app:summary tree construction}
\begin{algorithm}[H]
\caption{Hierarchical Summary Tree Construction}
\label{alg:summary_tree}
\begin{algorithmic}[1]

\Procedure{BuildSummaryTree}{$\mathcal{C}, E$}  
\Comment{$\mathcal{C} = \{C_1, C_2, \dots, C_m\}$ is the input set of text chunks, $E$ is the set of entities}

\State Initialize summary tree $\mathcal{S} \gets \emptyset$ 
\State Create root node $N_{\text{root}}$ with $\mathcal{C}_{\text{root}} = \mathcal{C}$ and entity set $E$, Generate global summary $S_{\text{root}}$
\State $\mathcal{S} \gets N_{\text{root}}$

\For{each $C_i \in \mathcal{C}$} 
    \State \Call{RecursiveProcess}{$C_i, N_{\text{root}}, E$}
\EndFor

\State \Return $\mathcal{S}$

\EndProcedure

\Function{RecursiveProcess}{$C, P, E$}  
    \Comment{$C$ is the current text, $P$ is its parent node, $E$ is the set of entities}
    \State Segment $C$ into meaningful sub-sections $\{C_1, C_2, \dots, C_k\}$ based on key entities
    \State Record segmentation boundaries $\{(s_i, e_i)\}$

    \For{each $C_i$ in $\{C_1, C_2, \dots, C_k\}$}
        \State Generate summary $S_i$ based on key entities in $E$
        \State Create child node $N_i$ with $(C_i, S_i)$
        \State Add $N_i$ to ${\mathit{Children}}(P)$
        \State \Call{RecursiveProcess}{$C_i, N_i, E$}
    \EndFor

\State \textbf{Stopping Conditions:}
\State 1. Stop if $\text{depth}(P) \geq D_{\max}$.
\State 2. Stop if no meaningful segmentation is found, as determined by the LLM.
\State 3. Stop if $\frac{|C_i|}{|C|} \geq \tau$ for some $C_i$.

\EndFunction

\end{algorithmic}
\end{algorithm}
\newpage
\subsection{Algorithm for Causal Graph Construction from Chunked Text}\label{app:causal_graph_construction}

\begin{algorithm}[htbp]
\caption{Causal Graph Construction from Chunked Text}
\begin{algorithmic}[1]
\Procedure{BuildGraphs}{$\mathcal{C}$}  
\Comment{$\mathcal{C}$ represents the set of all text chunks $\{C_1, \dots, C_k\}$}

\State Initialize $\mathcal{G} \gets \emptyset$ \Comment{Set of local graphs}
\For{each text chunk $C_i$}
    \State Extract entity set $V_i$ and causal edge set $E_i$
    \State $\mathcal{G} \gets \mathcal{G} \cup \{(V_i, E_i)\}$ 
    \Statex \quad \Comment{Store both entities and causal relationships}

    \EndFor
    \State \Return $\mathcal{G}$
\EndProcedure

\Procedure{MergeGlobalGraph}{$\mathcal{G}$}
    \State Initialize global edge set $E^* \gets \bigcup E_i, \forall E_i \in \mathcal{G}$
    \State Resolve conflicts in $E^*$ based on logical consistency
    \State \Return $E^*$
\EndProcedure
\end{algorithmic}
\end{algorithm}
\newpage
\section{Additional Experimental Results and Analysis}
\subsection{Effect of Top-k Entity Filtering}\label{app:topk}
\begin{table}[!ht]
    \centering
    \resizebox{\columnwidth}{!}{
    \begin{tabular}{llllll}
    
    \toprule
        Dataset & Win & Tie & Lose & W.R & N.W.R  \\ \midrule
        NarrativeQA & 2,278 & 30 & 764 & 67.15 & 35.75  \\ 
        QuALITY & 509 & 11 & 238 & 74.15 & 49.28  \\ 
        GOVREPORT & 381 & 2 & 236 & 61.55 & 23.42  \\ 
        MultiFieldQA-en & 161 & 1 & 92 & 63.39 & 27.17  \\ 
        MuSiQue & 247 & 0 & 151 & 62.06 & 24.12  \\ 
        QASPER & 218 & 7 & 123 & 62.64 & 27.30  \\ 
        QMSum & 279 & 2 & 119 & 69.75 & 40.00  \\ 
        MultiFieldQA-zh & 296 & 0 & 104 & 74.00 & 48.00  \\ 
        Avg & 298.71  & 3.29  & 151.86  & 66.84 & 34.38  \\ 
        \bottomrule
    \end{tabular}
    }
        \caption{Performance on Top-3 Filtered Entities Across Multiple Datasets}
\end{table}

\begin{table}[!ht]
    \centering
    \resizebox{\columnwidth}{!}{
    \begin{tabular}{llllll}
    \toprule
        Dataset & Win & Tie & Lose & W.R & N.W.R  \\ \midrule
        NarrativeQA & 2,386 & 23 & 665 & 66.54 & 34.51  \\ 
        QuALITY & 507 & 11 & 244 & 77.62 & 55.99  \\ 
        GOVREPORT & 361 & 4 & 252 & 58.51 & 17.67  \\ 
        MultiFieldQA-en & 160 & 1 & 91 & 63.49 & 27.38  \\ 
        MuSiQue & 270 & 2 & 128 & 67.50 & 35.50  \\ 
        QASPER & 232 & 3 & 123 & 64.80 & 30.45  \\ 
        QMSum & 287 & 3 & 110 & 71.75 & 44.25  \\ 
        MultiFieldQA-zh & 286 & 0 & 114 & 71.50 & 43.00  \\ 
        Avg & 300.43  & 3.43  & 151.71  & 67.71 & 36.09  \\ 
        \bottomrule
    \end{tabular}
    }
        \caption{Performance on Top-10 Filtered Entities Across Multiple Datasets}
\end{table}

\begin{table}[!ht]
    \centering
    \resizebox{\columnwidth}{!}{
    \begin{tabular}{llllll}
    \toprule
        Dataset & Win & Tie & Lose & W.R & N.W.R  \\ \midrule
         NarrativeQA & 2,280 & 31 & 766 & 63.85 & 29.16  \\ 
        QuALITY & 484 & 11 & 263 & 74.10 & 49.20  \\ 
        GOVREPORT & 379 & 1 & 240 & 61.13 & 22.42  \\ 
        MultiFieldQA-en & 158 & 1 & 87 & 64.23 & 28.86  \\ 
        MuSiQue & 252 & 2 & 146 & 63.00 & 26.50  \\ 
        QASPER & 193 & 3 & 156 & 54.83 & 10.51  \\ 
        QMSum & 259 & 2 & 138 & 64.91 & 30.33  \\ 
        MultiFieldQA-zh & 279 & 2 & 119 & 69.75 & 40.00  \\ 
        Avg & 286.29  & 3.14  & 164.14  & 64.47 & 29.62  \\ 
        \bottomrule
    \end{tabular}
    }
        \caption{Performance Across Multiple Datasets Without Top-k Filtering}
\end{table}
\subsection{Effect of Different Chunk Size}\label{app:chunk_size}
\begin{table}[H]
    \centering
    \resizebox{\columnwidth}{!}{
    \begin{tabular}{llllll}
    \toprule
        Dataset & Win & Tie & Lose & W.R & N.W.R  \\ \midrule
       NarrativeQA & 2,305 & 28 & 727 & 66.14 & 33.07  \\ 
        QuALITY & 500 & 6 & 250 & 75.33 & 51.57  \\ 
        GOVREPORT & 373 & 3 & 244 & 60.16 & 20.81  \\ 
        MultiFieldQA-en & 155 & 3 & 85 & 63.79 & 28.81  \\ 
        MuSiQue & 236 & 1 & 163 & 59.00 & 18.25  \\ 
        QASPER & 218 & 0 & 118 & 64.88 & 29.76  \\ 
        QMSum & 257 & 2 & 141 & 64.25 & 29.00  \\ 
        MultiFieldQA-zh & 328 & 1 & 67 & 82.83 & 65.91  \\ 
        Avg & 295.29  & 2.29  & 152.57  & 67.05 & 34.65  \\ 
        \bottomrule
    \end{tabular}
    }
        \caption{Performance Across Multiple Datasets under the 2048 Chunk Size Setting}
\end{table}

\begin{table}[H]
    \centering
    \resizebox{\columnwidth}{!}{
    \begin{tabular}{llllll}
    \toprule
        Dataset & Win & Tie & Lose & W.R & N.W.R  \\ \midrule
        NarrativeQA & 1,961 & 70 & 926 & 58.87 & 20.83  \\ 
        QuALITY & 438 & 23 & 283 & 66.32 & 35.00  \\ 
        GOVREPORT & 370 & 3 & 243 & 60.06 & 20.62  \\ 
        MultiFieldQA-en & 182 & 0 & 80 & 69.47 & 38.93  \\ 
        MuSiQue & 261 & 3 & 134 & 65.58 & 31.91  \\ 
        QASPER & 226 & 0 & 140 & 61.75 & 23.50  \\ 
        QMSum & 273 & 0 & 127 & 68.25 & 36.50  \\ 
        MultiFieldQA-zh & 327 & 1 & 72 & 81.75 & 63.75  \\ 
        Avg & 296.71  & 4.29  & 154.14  & 66.51 & 33.88  \\ 
        \bottomrule
    \end{tabular}
    }
        \caption{Performance Across Multiple Datasets under the 8192 Chunk Size Setting}
\end{table}

\begin{table}[H]
    \centering
    \resizebox{\columnwidth}{!}{
    \begin{tabular}{llllll}
    \toprule
        Dataset & Win & Tie & Lose & W.R & N.W.R  \\ \midrule
        NarrativeQA & 2,052 & 91 & 901 & 65.34 & 32.93  \\ 
        QuALITY & 492 & 17 & 244 & 67.41 & 37.81  \\ 
        GOVREPORT & 418 & 1 & 201 & 67.42 & 35.00  \\ 
        MultiFieldQA-en & 211 & 2 & 85 & 70.81 & 42.28  \\ 
        MuSiQue & 256 & 4 & 138 & 64.32 & 29.65  \\ 
        QASPER & 231 & 1 & 168 & 57.75 & 15.75  \\ 
        QMSum & 273 & 3 & 124 & 68.25 & 37.25  \\ 
        MultiFieldQA-zh & 334 & 0 & 66 & 83.50 & 67.00  \\ 
        Avg & 316.43  & 4.00  & 146.57  & 68.10 & 37.21  \\ 
        \bottomrule
    \end{tabular}
    }
        \caption{Performance Across Multiple Datasets without Chunking (Full Input)}
\end{table}

\begin{table}[H]
    \centering
    \resizebox{\columnwidth}{!}{
    \begin{tabular}{lcc}
    \toprule
        Model & w/length control & w/o length control  \\ \midrule
        GPT-4o & 66.56 & 64.38  \\ 
        GPT-4o-mini & 70.15 & 69.03  \\ 
        Deepseek-V3 & 72.71 & 65.3  \\ 
        QWQ-32B & 70.41 & 65.87  \\ 
        \bottomrule
    \end{tabular}
    
    }
    \caption{Comparison of evaluation scores with and without response length control.}
\end{table}\label{table:length_bias}
\subsection{Analyzing the Impact of Length Bias on Win Rate Evaluation}\label{app:length_bias}
To assess the potential influence of length bias—where evaluators or models may prefer longer responses—we analyze the win rate under a controlled setting. Specifically, we only consider response pairs where the length ratio between the shorter and longer response is no less than 0.8, ensuring comparable lengths and reducing the influence of response verbosity.

As shown in Table~\ref{table:length_bias}, the win rates under length-controlled evaluation are comparable to, and in some cases even higher than, those without length control. This suggests that our experimental results are robust and not significantly affected by length bias.

\section{Text Splitting Strategy}\label{app:text_split}

As an optional component in our framework, we adopt a chunking strategy to divide documents into manageable segments for effectively processing long-form textual inputs. There exist various approaches to splitting text, such as sentence-based, paragraph-based, and token-based methods. Each method offers trade-offs between semantic coherence and computational constraints. Moreover, chunking strategies can be adapted to different model context window sizes, enhancing their general applicability across various language models.

We utilize a token-aware splitting approach that recursively segments the input using structural cues (e.g., newlines and punctuation) while ensuring that the final chunk length adheres to a specified token limit. This method is particularly effective for maintaining coherence across chunks and avoiding abrupt cuts in meaning.

In future work, semantic similarity-based chunking methods may also be explored to further preserve semantic continuity and improve overall performance.

\section{Data Construction Details}\label{app:data_construction}

We construct our dataset by collecting and processing text samples from sources. Each source provides unique characteristics in terms of text length, domain, and accompanying metadata. Below we describe the key steps in our data construction process:

\subsection{Data Processing Pipeline}
\begin{itemize}
    \item \textbf{Text Truncation}: For English texts, we truncate by word count; for Chinese, we truncate by character count. The maximum length is adaptively set based on the dataset characteristics. For the Narrative QA dataset, the maximum length is set to 8192, while no specific length is set for other datasets.
    
    \item \textbf{Metadata Integration}: We carefully extract and format relevant metadata from each source to provide additional context for generating editing suggestions.
    
    \item \textbf{Deduplication}: We remove duplicate articles based on article IDs to ensure data uniqueness
    
    \item \textbf{Batch Processing}: For large datasets , we process data in batches to manage memory constraints.
    
    \item \textbf{Language Adaptation}: All prompts and processing logic are adapted to the target language (English or Chinese).
\end{itemize}

\subsection{Metadata Usage Across Datasets}
The metadata varies significantly across different datasets, providing diverse contextual information for the task. Table~\ref{tab:metadata} summarizes the key metadata components used from each dataset.

\begin{table*}[htbp]
\centering
\begin{tabular}{lll}
\hline
\textbf{Dataset} & \textbf{Primary Metadata Components} & \textbf{Purpose} \\
\hline
QuALITY & 
\begin{tabular}[c]{@{}l@{}}
- Source\\ 
- Author\\ 
- Topic\\ 
- QA pairs
\end{tabular} & 
\begin{tabular}[c]{@{}l@{}}
Provides provenance\\ 
and comprehension\\ 
questions
\end{tabular} \\
\hline
NarrativeQA & 
\begin{tabular}[c]{@{}l@{}}
- Summary\\ 
- Multiple QA pairs
\end{tabular} & 
\begin{tabular}[c]{@{}l@{}}
Offers narrative\\ 
understanding\\ 
and summaries
\end{tabular} \\
\hline
LVEval & 
\begin{tabular}[c]{@{}l@{}}
None (text only)
\end{tabular} & 
\begin{tabular}[c]{@{}l@{}}
Focuses on pure\\ 
text evaluation
\end{tabular} \\
\hline
LongBench & 
\begin{tabular}[c]{@{}l@{}}
- Questions\\ 
- Answers\\ 
- Summaries
\end{tabular} & 
\begin{tabular}[c]{@{}l@{}}
Provides task-specific\\ 
context
\end{tabular} \\
\hline
\end{tabular}
\caption{Meta-Information Usage in Different Datasets}
\end{table*}\label{tab:metadata}

The metadata serves multiple purposes in our editing task:
\begin{itemize}
    \item Providing background information about the text's origin and context
    \item Offering comprehension questions and answers that reveal key aspects of the text
    \item Supplying summaries that highlight main ideas
    \item Indicating the text's domain or topic.
\end{itemize}

For datasets without metadata, the modification suggestions are generated based solely on the text content, which uses the model's ability to analyze text without additional contextual clues.
\section{Criteria for Evaluating}\label{appendix:evaluation-criteria}
To evaluate the quality of the generated revisions, we design three key criteria, focusing on both fidelity to the original content and the naturalness of the edited text:

\begin{itemize}
    \item \textit{Faithfulness.} This refers to how well the revision reflects the intended edits without distorting the original meaning. The changes should be appropriately applied to the specified areas, while the rest of the text remains true to the original form.
    
    \item \textit{Logical Coherence.} This criterion examines whether the modified text remains logically connected throughout. Any local adjustment should be smoothly integrated into the full context, ensuring there are no contradictions or disjointed transitions.
    
    \item \textit{Fluency and Accessibility.} We assess how natural and smooth the revised text reads. The language should be polished and straightforward, allowing readers to easily grasp the content without unnecessary complexity or awkward phrasing.
\end{itemize}
We employ GPT-4o as the evaluation model to conduct comparisons based on the above criteria. To mitigate potential positional bias — where the order of presentation might influence the judgment — we perform evaluation twice for each comparison, swapping the positions of the two candidates. The final evaluation score is determined by aggregating the results from both positions, ensuring a more balanced and fair assessment.

As discussed in recent evaluations of LLM judges\cite{Zheng2023Judging}, existing models such as GPT-4\cite{OpenAI2023GPT} may exhibit several types of bias, including position bias, verbosity bias, self-enhancement bias, and limited capability in grading math and reasoning questions. In our evaluation framework, we mitigate these issues in the following ways. First, to address position bias, we exchange the order of the outputs and compute the win/lose relationship symmetrically, using the sum of both directions as the final score, thus ensuring position neutrality. Second, to assess verbosity bias, we control the length of generated responses and find no significant performance differences, indicating that longer answers do not receive disproportionate preference, as shown in Table~\ref{table:length_bias}. Third, to avoid self-enhancement bias, we use GPT-4o as our main evaluator, alongside three other models including GPT-4o itself, and observe no evidence of bias toward its own outputs. Lastly, our task focuses on open-ended generation rather than mathematical or logical reasoning, and therefore does not suffer from the grading limitations associated with such questions.

\end{document}